\documentclass{article} 
\usepackage{iclr2024_conference,times}

\usepackage{amsmath,amsfonts,bm}

\def\eqref#1{equation~\ref{#1}}

\def\1{\bm{1}}

\def\vs{{\bm{s}}}

\DeclareMathAlphabet{\mathsfit}{\encodingdefault}{\sfdefault}{m}{sl}
\SetMathAlphabet{\mathsfit}{bold}{\encodingdefault}{\sfdefault}{bx}{n}

\usepackage{times}
\usepackage{epsfig}
\usepackage{graphicx}
\usepackage{amsmath}
\usepackage{amssymb}
\usepackage{pifont}
\usepackage{booktabs}
\usepackage{arydshln}
\usepackage{color}
\usepackage{colortbl}
\usepackage{subcaption}
\usepackage{multirow}
\usepackage{float}
\usepackage{rotfloat}
\usepackage{capt-of}
\usepackage{longtable}
\usepackage{diagbox}
\usepackage{makecell}
\usepackage{pgfplots}
\usepackage{xspace}
\usepackage{wrapfig}
\usepackage{enumitem}

\pgfplotsset{compat=1.18}

\definecolor{linkc}{rgb}{0, 0.44, 0.74}
\definecolor{eqc}{rgb}{1, 0, 0}
\usepackage[pagebackref=false,breaklinks=true,colorlinks=True,urlcolor=eqc,citecolor=linkc,linkcolor=eqc,bookmarks=false]{hyperref}

\makeatletter
\DeclareRobustCommand\onedot{\futurelet\@let@token\@onedot}
\def\@onedot{\ifx\@let@token.\else.\null\fi\xspace}
\def\eg{\textit{e.g}\onedot}

\def\ie{\textit{i.e}\onedot}

\def\etc{\textit{etc}\onedot}
\def\vs{\textit{vs}\onedot}

\makeatother

\newcommand{\blfootnote}[1]{\begingroup
\renewcommand\thefootnote{}\footnote{#1}\addtocounter{footnote}{-1}
\endgroup}

\newcommand{\model}{\textsc{PixArt-$\alpha$}\xspace}

\newcommand{\COII}{CO\textsubscript{2}\xspace}
\definecolor{revise}{HTML}{0071BC}

\newcommand{\hl}[1]{\textcolor{black}{#1}}

\definecolor{mycolor_blue}{HTML}{E7EFFA}
\definecolor{mycolor_green}{HTML}{E6F8E0}
\definecolor{mycolor_gray}{HTML}{ECECEC}
\definecolor{pearDark}{HTML}{2980B9}

\title{\underline{\model}: Fast Training of Diffusion Transformer for Photorealistic Text-to-Image \\Synthesis}

\author{%
  Junsong Chen$^{1,2,3*}$,\space\space\space
  Jincheng Yu$^{1,4*}$,\space\space\space
  Chongjian Ge$^{1,3*}$,\space\space\space
  Lewei Yao$^{1,4*}$,\space\space\space
  Enze Xie$^{1\dagger}$,\space\space\space
  \\
  \textbf{Yue Wu$^{1}$,\space\space\space
  \vspace{0.15cm}
  Zhongdao Wang$^{1}$,\space\space\space
  James Kwok$^{4}$,\space\space\space
  Ping Luo$^{3}$,\space\space\space
  Huchuan Lu$^{2}$,\space\space\space
  Zhenguo Li$^{1}$\space\space\space
  }
  \\
  $^{1}$\normalfont{Huawei Noah's Ark Lab}\quad
  $^{2}$\normalfont{Dalian University of Technology}\quad
  $^{3}$\normalfont{HKU}\quad
  $^{4}$\normalfont{HKUST}\quad
  \vspace{0.15cm}\\
  \tt\small 
  {jschen@mail.dlut.edu.cn,
  rhettgee@connect.hku.hk,} \\
  \tt\small
  {\{yujincheng4,yao.lewei,xie.enze,Li.Zhenguo\}@huawei.com}
  \vspace{0.15cm}\\
  {\centering Project Page: \url{https://pixart-alpha.github.io/}}
}

\iclrfinalcopy 
\begin{document}
\maketitle

\blfootnote{$*$Equal contribution. Work done during the internships of the four students at Huawei Noah's Ark Lab.}
\blfootnote{$\dagger$ Project lead and corresponding author.}

\begin{abstract}
The most advanced text-to-image~(T2I) models require significant training costs ~(\eg, millions of GPU hours), seriously hindering the fundamental innovation for the AIGC community while increasing \COII emissions.  
This paper introduces \model, a Transformer-based T2I diffusion model whose image generation quality is competitive with state-of-the-art image generators~(\eg, Imagen, SDXL, and even Midjourney), reaching near-commercial application standards. Additionally, it supports high-resolution image synthesis up to 1024 $\times$ 1024 resolution with low training cost, as shown in Figure~\ref{fig:teaser} and \ref{fig:co2}.
To achieve this goal, three core designs are proposed:
(1) Training strategy decomposition: We devise three distinct training steps that respectively optimize pixel dependency, text-image alignment, and image aesthetic quality;
(2) Efficient T2I Transformer: We incorporate cross-attention modules into Diffusion Transformer~(DiT) to inject text conditions and streamline the computation-intensive class-condition branch;
(3) High-informative data:
We emphasize the significance of concept density in text-image pairs and leverage a large Vision-Language model to auto-label dense pseudo-captions to assist text-image alignment learning.
As a result, \model's training speed markedly surpasses existing large-scale T2I models, \eg, \model only takes \hl{12\%} of Stable Diffusion v1.5's training time~($\sim$\hl{753} \vs $\sim$6,250 A100 GPU days), saving nearly \$300,000~(\hl{\$28,400} \vs \$320,000) and reducing 90\% \COII emissions. 
Moreover, compared with a larger SOTA model, RAPHAEL, our training cost is merely 1\%.
Extensive experiments demonstrate that \model excels in image quality, artistry, and semantic control.
We hope \model will provide new insights to the AIGC community and startups to accelerate building their own high-quality yet low-cost generative models from scratch. 

\end{abstract}
\section{Introduction}

\begin{figure}[!ht]
\centering
\vspace{-1em}
\footnotesize
\includegraphics[width=0.95\linewidth]{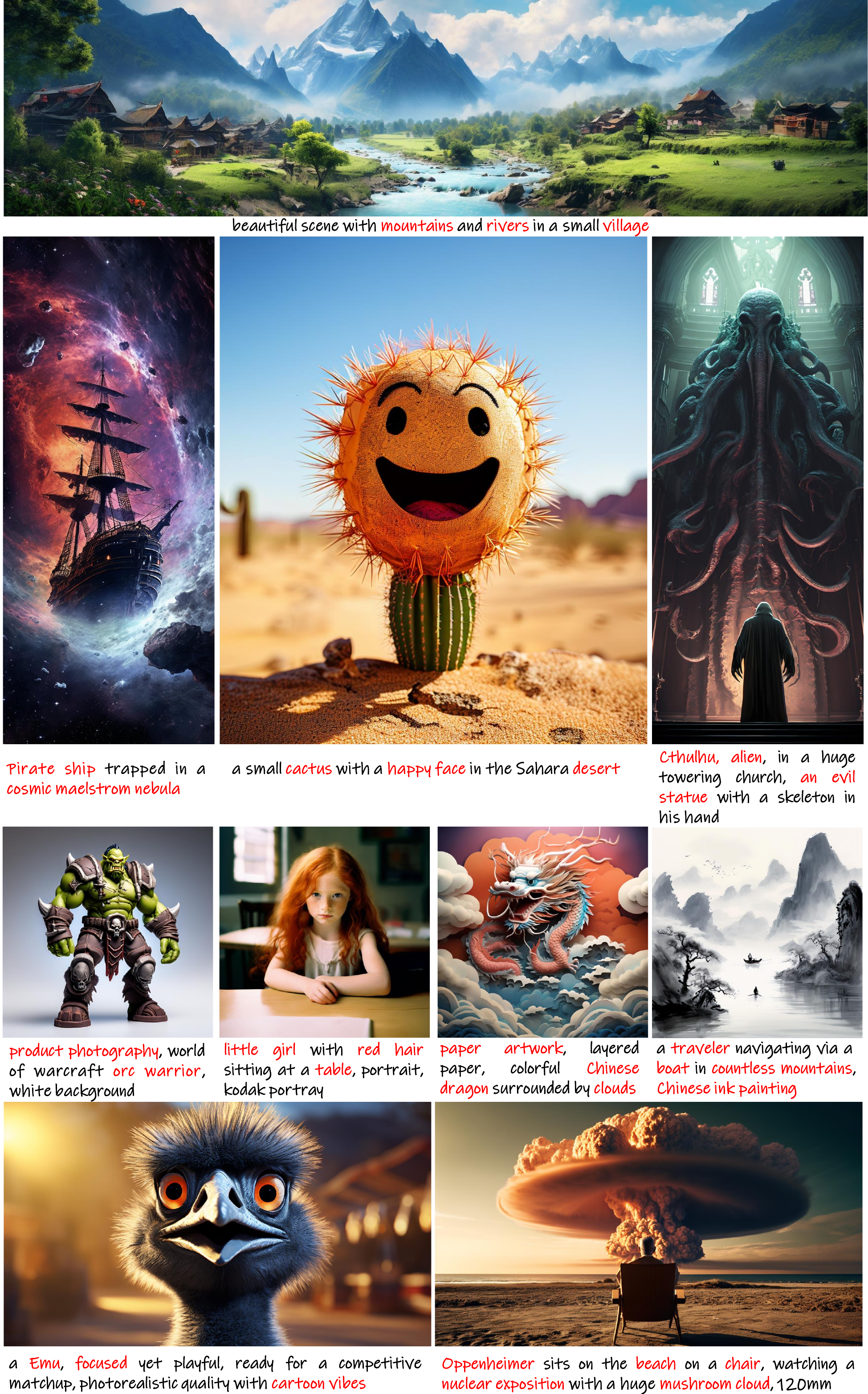}
\caption{Samples produced by \model exhibit exceptional quality, characterized by a remarkable level of fidelity and precision in adhering to the provided textual descriptions.}
\label{fig:teaser}
\end{figure}

Recently, the advancement of text-to-image~(T2I) generative models, such as DALL·E 2~\citep{Dalle-2}, Imagen~\citep{saharia2022photorealistic}, and Stable Diffusion~\citep{rombach2022high} has started a new era of photorealistic image synthesis, profoundly impacting numerous downstream applications, such as image editing~\citep{diffusionclip}, video generation~\citep{wu2022tuneavideo}, 3D assets creation~\citep{poole2022dreamfusion}, \etc.

\begin{figure}[htbp]
\vspace{-1em}
    \centering
    \includegraphics[width=\textwidth]{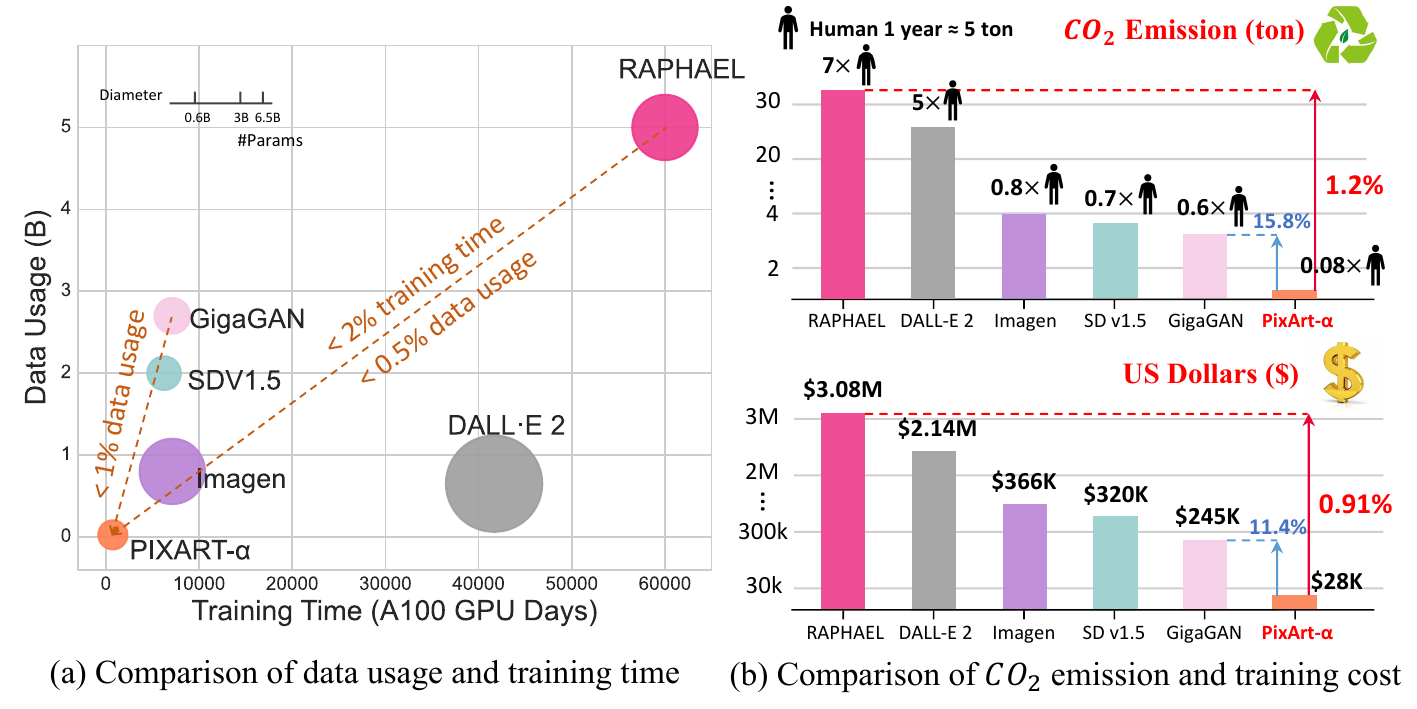}
    \caption{Comparisons of \COII emissions\protect\footnotemark{} and training cost\protect\footnotemark{} among T2I generators. \model achieves an exceptionally low training cost of \hl{\$28,400}.  Compared to RAPHAEL, our \COII emissions and training costs are merely \hl{1.2\%} and \hl{0.91\%}, respectively. 
    }
    \label{fig:co2}
    \vspace{-8pt}
\end{figure}

However, the training of these advanced models demands immense computational resources. For instance, training SDv1.5~\citep{podell2023sdxl} necessitates 6K A100 GPU days, approximately costing \$320,000, and the recent larger model,  RAPHAEL~\citep{xue2023raphael}, even costs 60K A100 GPU days -- requiring around \$3,080,000, as detailed in Table~\ref{tab:modelsize_params_mscocofid}.
Additionally, the training contributes substantial \COII emissions, posing environmental stress; \eg RAPHAEL's~\citep{xue2023raphael} training results in 35 tons of \COII emissions, equivalent to the amount one person emits over 7 years, as shown in Figure~\ref{fig:co2}. Such a huge cost imposes significant barriers for both the research community and entrepreneurs in accessing those models, causing a significant hindrance to the crucial advancement of the AIGC community.
Given these challenges, a pivotal question arises: 
{\it Can we develop a high-quality image generator with affordable resource consumption?}

\addtocounter{footnote}{-1}
\footnotetext{The method for estimating \COII emissions follows \cite{Luccioni2022co2}.} 
\addtocounter{footnote}{1}
\footnotetext{The training cost refers to the cloud GPU pricing from \cite{GPUprice} Azure in September 20, 2023.}

In this paper, we introduce \model, which significantly reduces computational demands of training while maintaining competitive image generation quality to the current state-of-the-art image generators, as illustrated in Figure~\ref{fig:teaser}. To achieve this, we propose three core designs:

\paragraph{Training strategy decomposition.}
We decompose the intricate text-to-image generation task into three streamlined subtasks: (1) learning the pixel distribution of natural images, (2) learning text-image alignment, and (3) enhancing the aesthetic quality of images.  
For the first subtask, we propose initializing the T2I model with a low-cost class-condition model, significantly reducing the learning cost. For the second and third subtasks, we formulate a training paradigm consisting of pretraining and fine-tuning: pretraining on text-image pair data rich in information density, followed by fine-tuning on data with superior aesthetic quality, boosting the training efficiency.

\paragraph{Efficient T2I Transformer.} Based on the Diffusion Transformer~(DiT)~\citep{Peebles2022DiT}, we incorporate cross-attention modules to inject text conditions and streamline the computation-intensive class-condition branch to improve efficiency.
Furthermore, we introduce a re-parameterization technique that allows the adjusted text-to-image model to load the original class-condition model's parameters directly. Consequently, we can leverage prior knowledge learned from ImageNet~\citep{deng2009imagenet} about natural image distribution to give a reasonable initialization for the T2I Transformer and accelerate its training.

\paragraph{High-informative data.} Our investigation reveals notable shortcomings in existing text-image pair datasets, exemplified by LAION~\citep{schuhmann2021laion}, where textual captions often suffer from a lack of informative content~(\ie, typically describing only a partial of objects in the images) and a severe long-tail effect~(\ie, with a large number of nouns appearing with extremely low frequencies). 
These deficiencies significantly hamper the training efficiency for T2I models and lead to millions of iterations to learn stable text-image alignments. 
To address them, we propose an auto-labeling pipeline utilizing the state-of-the-art vision-language model~(LLaVA~\citep{liu2023llava}) to generate captions on the SAM~\citep{kirillov2023segment}. Referencing in Section~\ref{sec:data}, the SAM dataset is advantageous due to its rich and diverse collection of objects, making it an ideal resource for creating high-information-density text-image pairs, more suitable for text-image alignment learning.

Our effective designs result in remarkable training efficiency for our model, costing only \hl{753} A100 GPU days and \hl{\$28,400}. As demonstrated in Figure~\ref{fig:co2}, our method consumes less than \hl{1.25\% training data volume compared to SDv1.5} and costs less than 2\% training time compared to RAPHAEL. Compared to RAPHAEL, our training costs are only 1\%, saving approximately \$3,000,000~(\model's \hl{\$28,400} \vs RAPHAEL's \$3,080,000).  Regarding generation quality, our user study experiments indicate that \model offers superior image quality and semantic alignment compared to existing SOTA T2I models~(\eg, DALL·E 2~\citep{Dalle-2}, Stable Diffusion~\citep{rombach2022high}, \etc), and its performance on T2I-CompBench~\citep{huang2023t2i} also evidences our advantage in semantic control. We hope our attempts to train T2I models efficiently can offer valuable insights for the AIGC community and help more individual researchers or startups create their own high-quality T2I models at lower costs.

\section{Method}
\subsection{Motivation}
The reasons for slow T2I training lie in two aspects: the training pipeline and the data. 

The T2I generation task can be decomposed into three aspects: {\bf Capturing Pixel Dependency:} Generating realistic images involves understanding intricate pixel-level dependencies within images and capturing their distribution; {\bf Alignment between Text and Image:} Precise alignment learning is required for understanding how to generate images that accurately match the text description; {\bf High Aesthetic Quality:} Besides faithful textual descriptions, being aesthetically pleasing is another vital attribute of generated images. Current methods entangle these three problems together and directly train from scratch using vast amount of data, resulting in inefficient training. 
To solve this issue, we disentangle these aspects into three stages, as will be described in Section~\ref{sec:train_stratedgy}.

\begin{figure}[ht]
    \centering
    \includegraphics[width=\textwidth]{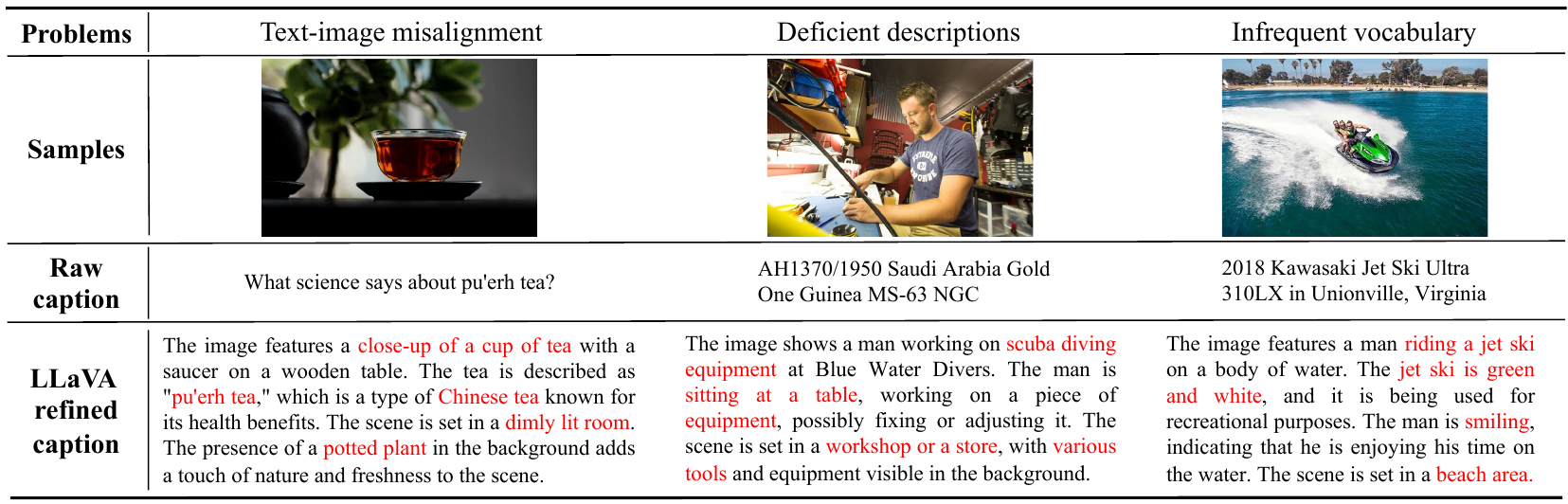}
    \caption{
    LAION raw captions \textit{v.s} LLaVA refined captions. LLaVA provides high-information-density captions that aid the model in grasping more concepts per iteration and boost text-image alignment efficiency.}    
    \label{fig:text-image-samples}
\vspace{-1mm}
\end{figure}

Another problem, depicted in Figure~\ref{fig:text-image-samples}, is with the quality of captions of the current dataset. The current text-image pairs often suffer from text-image misalignment, deficient descriptions, 
infrequent diverse vocabulary usage, and inclusion of low-quality data. These problems introduce difficulties in training, resulting in unnecessarily millions of iterations to achieve stable alignment between text and images. 
To address this challenge, we introduce an innovative auto-labeling pipeline to generate precise image captions, as will be described in Section~\ref{sec:data}.

\subsection{Training strategy Decomposition}\label{sec:train_stratedgy}
The model's generative capabilities can be gradually optimized by partitioning the training into three stages with different data types.

\textbf{Stage1: Pixel dependency learning.} The current class-guided approach~\citep{Peebles2022DiT} has shown exemplary performance in generating semantically coherent and reasonable pixels in individual images. Training a class-conditional image generation model~\citep{Peebles2022DiT} for natural images is relatively easy and inexpensive, as explained in Appendix~\ref{sec:pixel}. Additionally, we find that a suitable initialization can significantly boost training efficiency. Therefore, we boost our model from an ImageNet-pretrained model, and the architecture of our model is designed to be compatible with the pretrained weights.

\textbf{Stage2: Text-image alignment learning.} The primary challenge in transitioning from pretrained class-guided image generation to text-to-image generation is on how to achieve accurate alignment between significantly increased text concepts and images. 

 \begin{wrapfigure}{r}{0.5\textwidth}
  \centering
  \vspace{-2em}
  \includegraphics[width=0.48\textwidth]{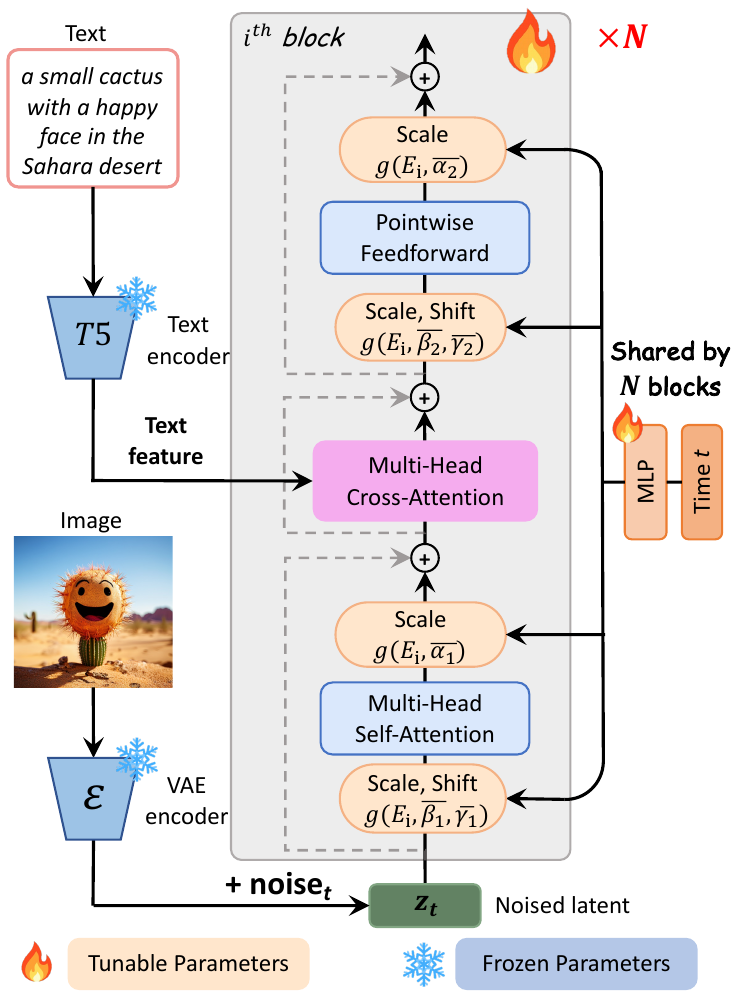}
  \caption{Model architecture of \model. A cross-attention module is integrated into each block to inject textual conditions. 
To optimize efficiency, all blocks share the same adaLN-single parameters for time conditions.}
  \vspace{-4em}
  \label{fig:pipeline-model}
\end{wrapfigure}

This alignment process is not only time-consuming but also inherently challenging. To efficiently facilitate this process, we construct a dataset consisting of precise text-image pairs with high concept density. The data creation pipeline will be described in Section~\ref{sec:data}. By employing accurate and information-rich data, our training process can efficiently handle a larger number of nouns in each iteration while encountering considerably less ambiguity compared to previous datasets. This strategic approach empowers our network to align textual descriptions with images effectively.

\textbf{Stage3: High-resolution and aesthetic image generation.} In the third stage, we fine-tune our model using high-quality aesthetic data for high-resolution image generation. 
Remarkably, we observe that the adaptation process in this stage converges significantly faster, primarily owing to the strong prior knowledge established in the preceding stages.

Decoupling the training process into different stages significantly alleviates the training difficulties and achieves highly efficient training.

\subsection{Efficient T2I Transformer}\label{sec:architecture}

\model adopts the Diffusion Transformer (DiT)~\citep{Peebles2022DiT} as the base architecture and innovatively 
tailors the Transformer blocks to handle the unique challenges of T2I tasks, as depicted in Figure~\ref{fig:pipeline-model}. Several dedicated designs are proposed as follows:

\begin{itemize}[leftmargin=1em]
\item {\it Cross-Attention layer.} We incorporate a multi-head cross-attention layer to the DiT block. It is positioned between the self-attention layer and feed-forward layer so that the model can flexibly interact with the text embedding extracted from the language model. To facilitate the pretrained weights, we initialize the output projection layer in the cross-attention layer to zero, effectively acting as an identity mapping and preserving the input for the subsequent layers.

\item {\it AdaLN-single.} We find that the linear projections in the adaptive normalization layers~\citep{perez2018film} ({\it adaLN}) module of the DiT account for a substantial proportion (27\%) of the parameters. Such a large number of parameters is not useful since the class condition is not employed for our T2I model.
Thus, we propose adaLN-single, which only uses time embedding as input in the first block for independent control (shown on the right side of Figure~\ref{fig:pipeline-model}). Specifically, in the $i$th block, let $S^{(i)}=[\beta^{(i)}_1, \beta^{(i)}_2, \gamma^{(i)}_1, \gamma^{(i)}_2, \alpha^{(i)}_1, \alpha^{(i)}_2]$ be a tuple of all the scales and shift parameters in {\it adaLN}.
In the DiT, $S^{(i)}$ is obtained through a block-specific MLP $S^{(i)} = f^{(i)}(c+t)$, where $c$ and $t$ denotes the class condition and time embedding, respectively. However, in adaLN-single, one global set of shifts and scales are computed as $\overline{S} = f(t)$ only at the first block which is shared across all the blocks. 
Then, $S^{(i)}$ is obtained as $S^{(i)} = g(\overline{S}, E^{(i)})$, where $g$ is a summation function, and $E^{(i)}$ is a layer-specific trainable embedding with the same shape as $\overline{S}$, which adaptively adjusts the scale and shift parameters in different blocks. 

\item {\it Re-parameterization.} To utilize the aforementioned pretrained weights, all $E^{(i)}$'s are initialized to values that yield the same $S^{(i)}$ as the DiT without $c$ for a selected $t$ (empirically, we use $t=500$). This design effectively replaces the layer-specific MLPs with a global MLP and layer-specific trainable embeddings while preserving compatibility with the pretrained weights.
\end{itemize}

Experiments demonstrate that incorporating a global MLP and layer-wise embeddings for time-step information, as well as cross-attention layers for handling textual information, persists the model's generative abilities while effectively reducing its size.

\subsection{Dataset construction}\label{sec:data}
\paragraph{Image-text pair auto-labeling.}
The captions of the LAION dataset exhibit various issues, such as text-image misalignment, deficient descriptions, and infrequent vocabulary as shown in Figure~\ref{fig:text-image-samples}. To generate captions with high information density, we leverage the state-of-the-art vision-language model LLaVA~\citep{liu2023llava}. Employing the prompt, ``\textit{Describe this image and its style in a very detailed manner}'', we have significantly improved the quality of captions, as shown in Figure~\ref{fig:text-image-samples}.

However, it is worth noting that the LAION dataset predominantly comprises of simplistic product previews from shopping websites, which are not ideal for training text-to-image generation that seeks diversity in object combinations. Consequently, we have opted to utilize the SAM dataset~\citep{kirillov2023segment}, which is originally used for segmentation tasks but features imagery rich in diverse objects. By applying LLaVA to SAM, we have successfully acquired high-quality text-image pairs characterized by a high concept density, as shown in Figure~\ref{fig:llava-dialog} and Figure~\ref{fig:text-image-samples-sam} in the Appendix.

In the third stage, we construct our training dataset by incorporating JourneyDB~\citep{pan2023journeydb} and a 10M internal dataset to enhance the aesthetic quality of generated images beyond realistic photographs. Refer to Appendix~\ref{sec:pixel} for details.

\begin{wraptable}{r}{8cm}
\centering
    \vspace{-1em}
    \caption{Statistics of noun concepts for different datasets. {\bf VN}: valid distinct nouns (appearing more than 10 times); {\bf DN}: total distinct nouns; {\bf Average}: average noun count per image.
    } 
\label{tab:laion_sam_llava_compare}
\resizebox{\linewidth}{!}{ 
\begin{tabular}{cccccc}
\toprule
\bf Dataset & \bf VN/DN  & \bf Total Noun & \bf Average \\
\midrule
LAION  & 210K/2461K = 8.5\% & 72.0M & 6.4/Img \\
LAION-LLaVA & 85K/646K = 13.3\% & 233.9M & 20.9/Img \\
SAM-LLaVA & 23K/124K = 18.6\% & 327.9M & 29.3/Img \\
Internal & 152K/582K = 26.1\% & 136.6M & 12.2/Img \\
\bottomrule

\end{tabular}
}
\vspace{-1em}
\end{wraptable}

As a result, we show the vocabulary analysis~\citep{Nouncount} in Table~\ref{tab:laion_sam_llava_compare}, and we define the valid distinct nouns as those appearing more than 10 times in the dataset. We apply LLaVA on LAION to generate LAION-LLaVA. 
The LAION dataset has 2.46 M distinct nouns, but only 8.5\% are valid. This valid noun proportion significantly increases from 8.5\% to 13.3\% with LLaVA-labeled captions. 
Despite LAION's original captions containing a staggering 210K distinct nouns, its total noun number is a mere 72M. However, LAION-LLaVA contains 234M noun numbers with 85K distinct nouns, and the average number of nouns per image increases from 6.4 to 21, indicating the incompleteness of the original LAION captions. Additionally, SAM-LLaVA outperforms LAION-LLaVA with a total noun number of 328M and 30 nouns per image, demonstrating SAM contains richer objectives and superior informative density per image. Lastly, the internal data also ensures sufficient valid nouns and average information density for fine-tuning. LLaVA-labeled captions significantly increase the valid ratio and average noun count per image, improving concept density.

\section{Experiment}
This section begins by outlining the detailed training and evaluation protocols. Subsequently, we provide comprehensive comparisons across three main metrics. We then delve into the critical designs implemented in \model to achieve superior efficiency and effectiveness through ablation studies. Finally, we demonstrate the versatility of our \model through application extensions.

\subsection{Implementation details}
{\bf Training Details.} We follow Imagen~\citep{saharia2022photorealistic} and DeepFloyd~\citep{deepfloyd} to employ the T5 large language model~(\ie, 4.3B Flan-T5-XXL) as the text encoder for conditional feature extraction, and use DiT-XL/2~\citep{Peebles2022DiT} as our base network architecture. Unlike previous works that extract a standard and fixed 77 text tokens, we adjust the length of extracted text tokens to 120, as the caption curated in \model is much denser to provide more fine-grained details. To capture the latent features of input images, we employ a pre-trained and frozen VAE from LDM~\citep{rombach2022high}. Before feeding the images into the VAE, we resize and center-crop them to have the same size. We also employ multi-aspect augmentation introduced in SDXL~\citep{podell2023sdxl} to enable arbitrary aspect image generation. 
The AdamW optimizer~\citep{loshchilov2017adamw} is utilized with a weight decay of 0.03 and a constant 2e-5 learning rate. 
Our final model is trained on 64 V100 for approximately \hl{26} days.
See more details in Appendix~\ref{sec:pixel}.

{\bf Evaluation Metrics.}  We comprehensively evaluate \model via three primary metrics, \ie, Fréchet Inception Distance~(FID)~\citep{heusel2017gans} on MSCOCO dataset~\citep{microsoftcoco}, compositionality on T2I-CompBench~\citep{huang2023t2i}, and human-preference rate on user study.

\subsection{Performance Comparisons and Analysis}
{\bf Fidelity Assessment.} 
The FID is a metric to evaluate the quality of generated images. The comparison between our method and other methods in terms of FID and their training time is summarized in Table~\ref{tab:modelsize_params_mscocofid}. When tested for zero-shot performance on the COCO dataset, \model achieves a FID score of 7.32. It is particularly notable as it is accomplished in merely 12\% of the training time~(\hl{753} \vs 6250 A100 GPU days) and merely \hl{1.25\%} of the training samples~(25M \vs \hl{2B} images) relative to the second most efficient method. 
Compared to state-of-the-art methods typically trained using substantial resources, \model remarkably consumes approximately 2\% of the training resources while achieving a comparable FID performance. Although the best-performing model (RAPHEAL) exhibits a lower FID, it relies on unaffordable resources (\ie, $200\times$ more training samples, \hl{$80\times$} longer training time, and $5\times$ more network parameters than \model). We argue that FID may not be an appropriate metric for image quality evaluation, and it is more appropriate to use the evaluation of human users, as stated in Appendix~\ref{appendix:fid}. We leave scaling of \model{} for future exploration for performance enhancement.

{\bf Alignment Assessment.} 
Beyond the above evaluation, we also assess the alignment between the generated images and text condition using T2I-Compbench~\citep{huang2023t2i}, a comprehensive benchmark for evaluating the compositional text-to-image generation capability. As depicted in Table~\ref{benchmark:t2icompbench}, we evaluate several crucial aspects, including attribute binding, object relationships, and complex compositions. \model{} exhibited outstanding performance across nearly all~(5/6) evaluation metrics. This remarkable performance is primarily attributed to the text-image alignment learning in Stage 2 training described in Section~\ref{sec:train_stratedgy}, where high-quality text-image pairs were leveraged to achieve superior alignment capabilities.


\begin{table}[t]
\vspace{-3em}
\centering
\caption{We thoroughly compare the \model with recent T2I models, considering several essential factors: model size, the total volume of training images, COCO FID-30K scores~(zero-shot), and the computational cost~(GPU days\protect\footnotemark{}). Our highly effective approach significantly reduces resource consumption, including training data usage and training time. The baseline data is sourced from GigaGAN~\citep{kang2023gigagan}. \hl{`+' in the table denotes an unknown internal dataset size.}}
\label{tab:modelsize_params_mscocofid}
\resizebox{0.75\linewidth}{!}{ 

\begin{tabular}{lccccc}
        \toprule
    	\bf Method & \bf Type & \bf $\mathrm{\#}$Params & \bf $\mathrm{\#}$Images  & \bf FID-30K$\downarrow$ & \bf GPU days \\
    	\midrule
    	DALL·E & Diff   & 12.0B & \hl{250M} & 27.50 & - \\
            GLIDE  & Diff   & 5.0B  & \hl{250M} & 12.24 & - \\
            LDM    & Diff   & 1.4B  & \hl{400M} & 12.64 & -\\
            DALL·E 2 & Diff & 6.5B  & \hl{650M} & 10.39 & 41,667 A100\\
            SDv1.5 & Diff   & 0.9B  & \hl{2000M} & 9.62  & 6,250 A100\\
            GigaGAN & GAN   & 0.9B  & \hl{2700M} & 9.09  & 4,783 A100 \\
            Imagen & Diff   & 3.0B  & \hl{860M} & 7.27  & 7,132 A100\\
            RAPHAEL & Diff  & 3.0B  & \hl{5000M+} & 6.61  & 60,000 A100 \\
            \midrule
    	\model & Diff   & \cellcolor{pearDark!20}{0.6B} & \cellcolor{pearDark!20}{\hl{25M}} & 7.32 & \cellcolor{pearDark!20}{753 A100} \\
        \bottomrule

\end{tabular}
\vspace{-3mm}
}
\end{table}

\footnotetext{To ensure fairness, we convert the V100 GPU days~(\hl{1656}) of our training to A100 GPU days~(\hl{753}), assuming a 2.2$\times$ speedup in U-Net training on A100 compared to V100, or equivalent to 332 A100 GPU days with a 5$\times$ speedup in Transformer training, as per~\citet{rombach2022high,A100vsV100}.} 
 
\begin{table}[ht]
\centering
\caption{Alignment evaluation on T2I-CompBench. 
\model demonstrated exceptional performance in attribute binding, object relationships, and complex compositions, indicating our method achieves superior compositional generation ability.
We highlight the best value in \colorbox{pearDark!20}{blue}, and the second-best value in \colorbox{mycolor_green}{green}. The baseline data are sourced from ~\cite{huang2023t2i}.} 
\label{benchmark:t2icompbench}
\resizebox{0.9\linewidth}{!}{ 
\begin{tabular}
{lcccccc}
\toprule
\multicolumn{1}{c}
{\multirow{2}{*}{\bf Model}} & \multicolumn{3}{c}{\bf Attribute Binding } & \multicolumn{2}{c}{\bf Object Relationship} & \multirow{2}{*}{\bf Complex$\uparrow$}
\\
\cmidrule(lr){2-4}\cmidrule(lr){5-6}

&
{\bf Color $\uparrow$ } &
{\bf Shape$\uparrow$} &
{\bf Texture$\uparrow$} &
{\bf Spatial$\uparrow$} &
{\bf Non-Spatial$\uparrow$} &
\\
\midrule
Stable v1.4  & 0.3765 & 0.3576 & 0.4156 & 0.1246 & 0.3079 & 0.3080  \\
Stable v2 & 0.5065 & 0.4221 & 0.4922 & 0.1342 & 0.3096 & 0.3386  \\
Composable v2 & 0.4063 & 0.3299 & 0.3645 & 0.0800 & 0.2980 & 0.2898  \\
Structured v2 & 0.4990 & 0.4218 & 0.4900 & 0.1386 & 0.3111 & 0.3355  \\
Attn-Exct v2 & 0.6400 & 0.4517 & 0.5963 & 0.1455 & 0.3109 & 0.3401  \\
GORS  & \cellcolor{mycolor_green}{0.6603} & 0.4785 & 0.6287 & 0.1815 & \cellcolor{pearDark!20}{0.3193} & 0.3328  \\
Dalle-2 & 0.5750 & \cellcolor{mycolor_green}{0.5464} & \cellcolor{mycolor_green}{0.6374} & 0.1283 & 0.3043 & 0.3696  \\
SDXL & 0.6369 & 0.5408 & 0.5637 & \cellcolor{mycolor_green}{0.2032} & 0.3110 & \cellcolor{mycolor_green}{0.4091}  \\
\midrule

\model~ & \cellcolor{pearDark!20}{0.6886} & \cellcolor{pearDark!20}{0.5582} & \cellcolor{pearDark!20}{0.7044} & \cellcolor{pearDark!20}{0.2082} & \cellcolor{mycolor_green}{0.3179} & \cellcolor{pearDark!20}{0.4117}  \\
\bottomrule
\end{tabular}
}
\vspace{-1em}
\end{table}

{\bf User Study.} While quantitative evaluation metrics measure the overall distribution of two image sets, they may not comprehensively evaluate the visual quality of the images. 
Consequently, we conducted a user study to supplement our evaluation and provide a more intuitive assessment of \model{}'s performance.
Since user study involves human evaluators and can be time-consuming, we selected the top-performing models, namely DALLE-2, SDv2, SDXL, and DeepFloyd, which are accessible through APIs and capable of generating images.

\begin{figure}[ht]
\begin{center}
\footnotesize
\includegraphics[width=1\linewidth]{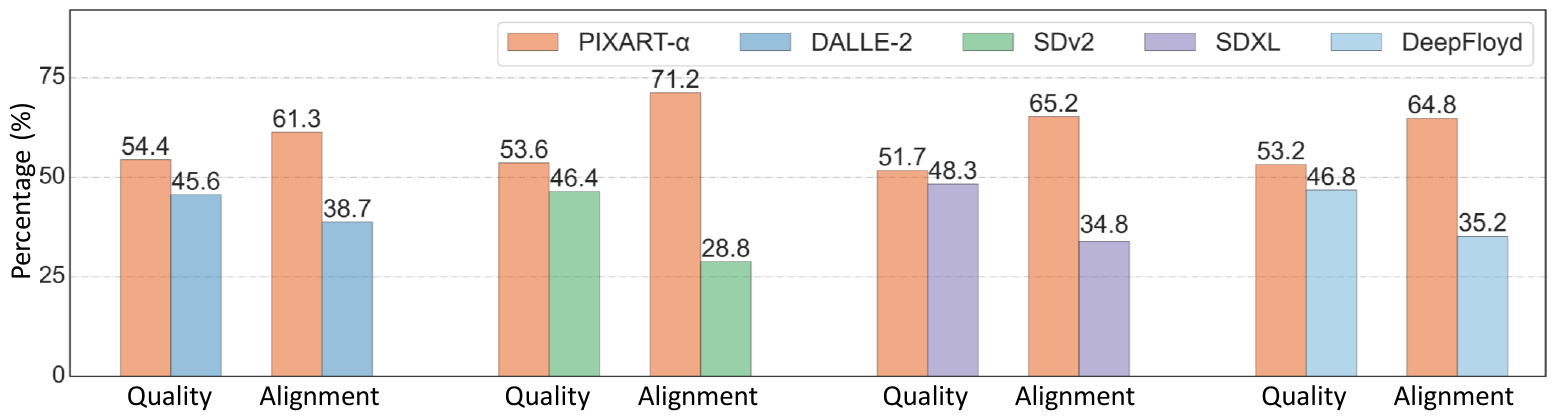}
\caption{User study on 300 fixed prompts from \cite{feng2023ernie}. The ratio values indicate the percentages of participants preferring the corresponding model. \model achieves a superior performance in both quality and alignment.} 
\label{fig:user_study}
\end{center} 
\vspace{-1em}
\end{figure}

For each model, we employ a consistent set of 300 prompts from~\cite{feng2023ernie} to generate images.
These images are then distributed among 50 individuals for evaluation. Participants are asked to rank each model based on the perceptual quality of the generated images and the precision of alignments between the text prompts and the corresponding images. The results presented in Figure~\ref{fig:user_study} clearly indicate that \model{} excels in both higher fidelity and superior alignment.
For example, compared to SDv2, a current top-tier T2I model, \model exhibits a 7.2\% improvement in image quality and a substantial 42.4\% enhancement in alignment.

\begin{figure}[ht]
\vspace{-1em}
\begin{center}
\footnotesize
\includegraphics[width=1\linewidth]{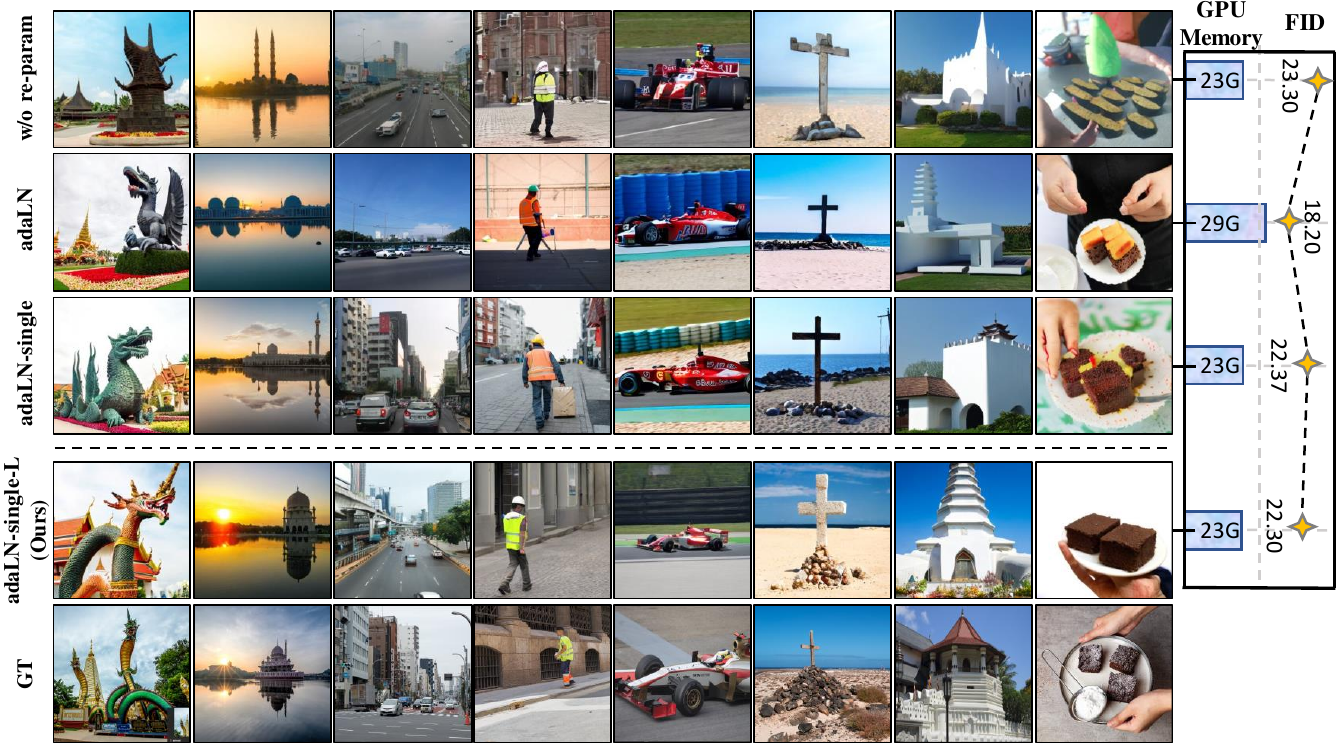}
\caption{\textbf{Left}: Visual comparison of ablation studies are presented.
\textbf{Right}: Zero-shot FID-2K on SAM, and GPU memory usage. Our method is on par with the ``{\it adaLN}'' and saves 21\% in GPU memory.
Better zoom in 200\%.}
\label{fig:ablation-study}
\end{center}
\vspace{-1.5em}
\end{figure}

\subsection{Ablation Study}
We then conduct ablation studies on the crucial modifications discussed in Section~\ref{sec:architecture}, including structure modifications and re-parameterization design. In Figure~\ref{fig:ablation-study}, we provide visual results and perform a FID analysis. We randomly choose 8 prompts from the SAM test set for visualization and compute the zero-shot FID-5K score on the SAM dataset. Details are described below.

``{\it w/o re-param}'' results are generated from the model trained from scratch without re-parameterization design. We supplemented with an additional 200K iterations to compensate for the missing iterations from the pretraining stage for a fair comparison.
``{\it adaLN}'' results are from the model following the DiT structure to use the sum of time and text feature as input to the MLP layer for the scale and shift parameters within each block. 
``{\it adaLN-single}'' results are obtained from the model using Transformer blocks with the adaLN-single module in Section~\ref{sec:architecture}. In both ``{\it adaLN}'' and ``{\it adaLN-single}'', we employ the re-parameterization design and training for 200K iterations.

As depicted in Figure~\ref{fig:ablation-study}, despite ``{\it adaLN}'' performing lower FID, its visual results are on par with our ``{\it adaLN-single}'' design. 
The GPU memory consumption of ``{\it adaLN}'' is 29GB, whereas ``{\it adaLN-single}'' achieves a reduction to 23GB, saving 21\% in GPU memory consumption. Furthermore, considering the model parameters, the ``{\it adaLN}'' method consumes 833M, whereas our approach reduces to a mere 611M, resulting in an impressive 26\% reduction.
``{\it adaLN-single-L (Ours)}'' results are generated from the model with same setting as ``{\it adaLN-single}'', but training for a \textbf{L}onger training period of 1500K iterations.
Considering memory and parameter efficiency, we incorporate the ``{\it adaLN-single-L}'' into our final design.

The visual results clearly indicate that, although the differences in FID scores between the ``{\it adaLN}'' and ``{\it adaLN-single}'' models are relatively small, a significant discrepancy exists in their visual outcomes. The ``{\it w/o re-param}'' model consistently displays distorted target images and lacks crucial details across the entire test set.

\section{Related work}\label{related_work_overview}
We review related works in three aspects: Denoising diffusion probabilistic models~(DDPM), Latent Diffusion Model, and Diffusion Transformer. More related works can be found in Appendix~\ref{appendix:related}. DDPMs~\citep{ho2020denoising,sohl2015deep} have emerged as highly successful approaches for image generation, which employs an iterative denoising process to transform Gaussian noise into an image. Latent Diffusion Model~\citep{rombach2022high} enhances the traditional DDPMs by employing score-matching on the image latent space and introducing cross-attention-based controlling. Witnessed the success of Transformer architecture on many computer vision tasks, Diffusion Transformer~(DiT)~\citep{Peebles2022DiT} and its variant~\citep{bao2023all,zheng2023fast} further replace the Convolutional-based U-Net~\citep{ronneberger2015u} backbone with Transformers for increased scalability.
\section{Conclusion}
In this paper, we introduced \model, a Transformer-based text-to-image (T2I) diffusion model, which achieves superior image generation quality while significantly reducing training costs and \COII emissions. Our three core designs, including the training strategy decomposition, efficient T2I Transformer and high-informative data, contribute to the success of \model.
Through extensive experiments, we have demonstrated that \model achieves near-commercial application standards in image generation quality. 
With the above designs, \model provides new insights to the AIGC community and startups, enabling them to build their own high-quality yet low-cost T2I models. We hope that our work inspires further innovation and advancements in this field.
\paragraph{Acknowledgement.} We would like to express our gratitude to Shuchen Xue for identifying and correcting the FID score in the paper.
\newpage

\clearpage
\appendix


\newpage
\section{Appendix}
\subsection{Related work}\label{appendix:related}

\subsubsection{Denoising diffusion probabilistic models}
Diffusion models~\citep{ho2020denoising,sohl2015deep} and score-based generative models~\citep{song2019generative,song2020score} have emerged as highly successful approaches for image generation, surpassing previous generative models such as GANs~\citep{goodfellow2014generative}, VAEs~\citep{kingma2013auto}, and Flow~\citep{rezende2015variational}. Unlike traditional models that directly map from a Gaussian distribution to the data distribution, diffusion models employ an iterative denoising process to transform Gaussian noise into an image that follows the data distribution. This process can be reversely learned from an untrainable forward process, where a small amount of Gaussian noise is iteratively added to the original image. 

\subsubsection{Latent Diffusion Model}
Latent Diffusion Model (\textit{a.k.a.} Stable diffusion)~\citep{rombach2022high} is a recent advancement in diffusion models. This approach enhances the traditional diffusion model by employing score-matching on the image latent space and introducing cross-attention-based controlling. The results obtained with this approach have been impressive, particularly in tasks involving high-density image generation, such as text-to-image synthesis. This has served as a source of inspiration for numerous subsequent works aimed at improving text-to-image synthesis, including those by~\citet{saharia2022photorealistic,balaji2022ediffi,feng2023ernie,xue2023raphael,podell2023sdxl}, and others. Additionally, Stable diffusion and its variants have been effectively combined with various low-cost fine-tuning~\citep{hu2021lora, xie2023difffit} and customization~\citep{zhang2023adding, mou2023t2i} technologies.

\subsubsection{Diffusion Transformer}
Transformer architecture~\citep{vaswani2017attention} have achieved great success in language models~\citep{radford2018improving,radford2019language}, and many recent works~\citep{dosovitskiy2020image,he2022masked} show it is also a promising architecture on many computer vision tasks like image classification~\citep{touvron2021training,zhou2021deepvit,yuan2021tokens,han2021transformer}, object detection~\citep{liu2021swin,wang2021pyramid,wang2022pvt, ge2023metabev, carion2020end}, semantic segmentation~\citep{zheng2021rethinking,xie2021segformer,strudel2021segmenter} and so on~\citep{sun2020transtrack,li2022panoptic,zhao2021point,liu2022video,he2022masked,li2022bevformer}. 
The Diffusion Transformer (DiT)~\citep{Peebles2022DiT} and its variant~\citep{bao2023all,zheng2023fast} follow the step to further replace the Convolutional-based U-Net~\citep{ronneberger2015u} backbone with Transformers. This architectural choice brings about increased scalability compared to U-Net-based diffusion models, allowing for the straightforward expansion of its parameters.
In our paper, we leverage DiT as a scalable foundational model and adapt it for text-to-image generation tasks.

\subsection{\model{} \vs Midjourney}\label{sec:compare-mj}
\begin{figure}[ht]
\begin{center}
\footnotesize
\includegraphics[width=1\linewidth]{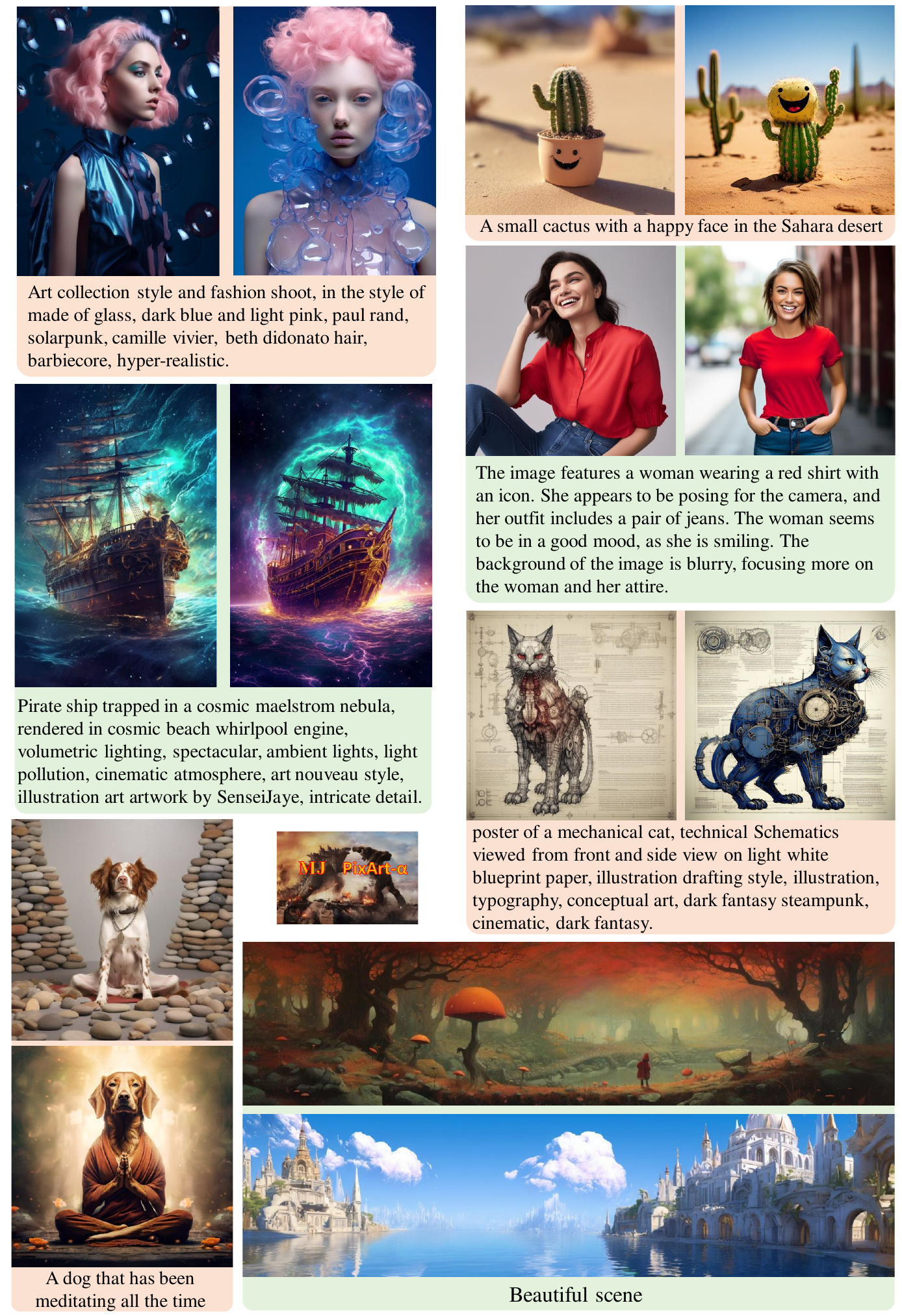}
\caption{Comparisons with Midjourney. The prompts used here are randomly sampled online. To ensure a fair comparison, we select the first result generated by both models.
\textbf{\textit{We encourage readers to guess which image corresponds to Midjourney and which corresponds to \model. The answer is revealed at the end of the paper.}}
}
\label{fig:compare_mj}
\end{center} 
\end{figure}

In Figure~\ref{fig:compare_mj}, we present the images generated using \model{} and the current SOTA product-level method Midjourney~\citep{Midjourney} with randomly sampled prompts online. Here, we conceal the annotations of images belonging to which method. Readers are encouraged to make assessments based on the prompts provided. The answers will be disclosed at the end of the appendix. 

\subsection{\model \vs Prestigious Diffusion Models}
\begin{figure}[ht]
\begin{center}
\footnotesize
\includegraphics[width=1\linewidth]{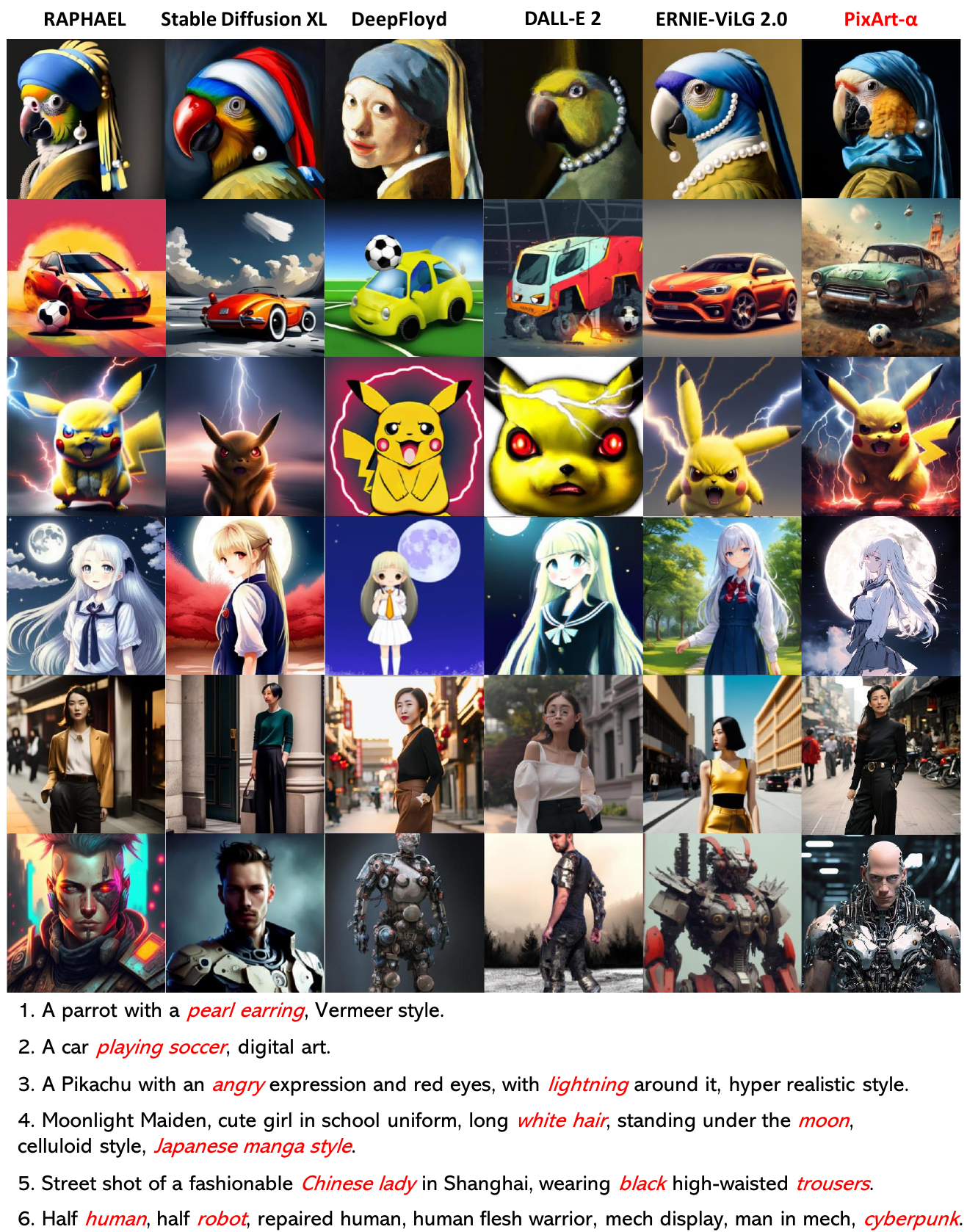}
\caption{\textbf{Comparisons} of \model with recent representative generators, Stable Diffusion XL, DeepFloyd, DALL-E 2, ERNIE-ViLG 2.0, and RAPHAEL. They are given the same prompts as in RAPHAEL\citep{xue2023raphael}, where the words that the human artists yearn to preserve within the generated images are highlighted in red. The specific prompts for each row are provided at the bottom of the figure. Better zoom in 200\%.}
\label{fig:comparefig_multiple_methods}
\end{center} 
\end{figure}

\begin{figure}[ht]
\begin{center}
\footnotesize
\includegraphics[width=1\linewidth]{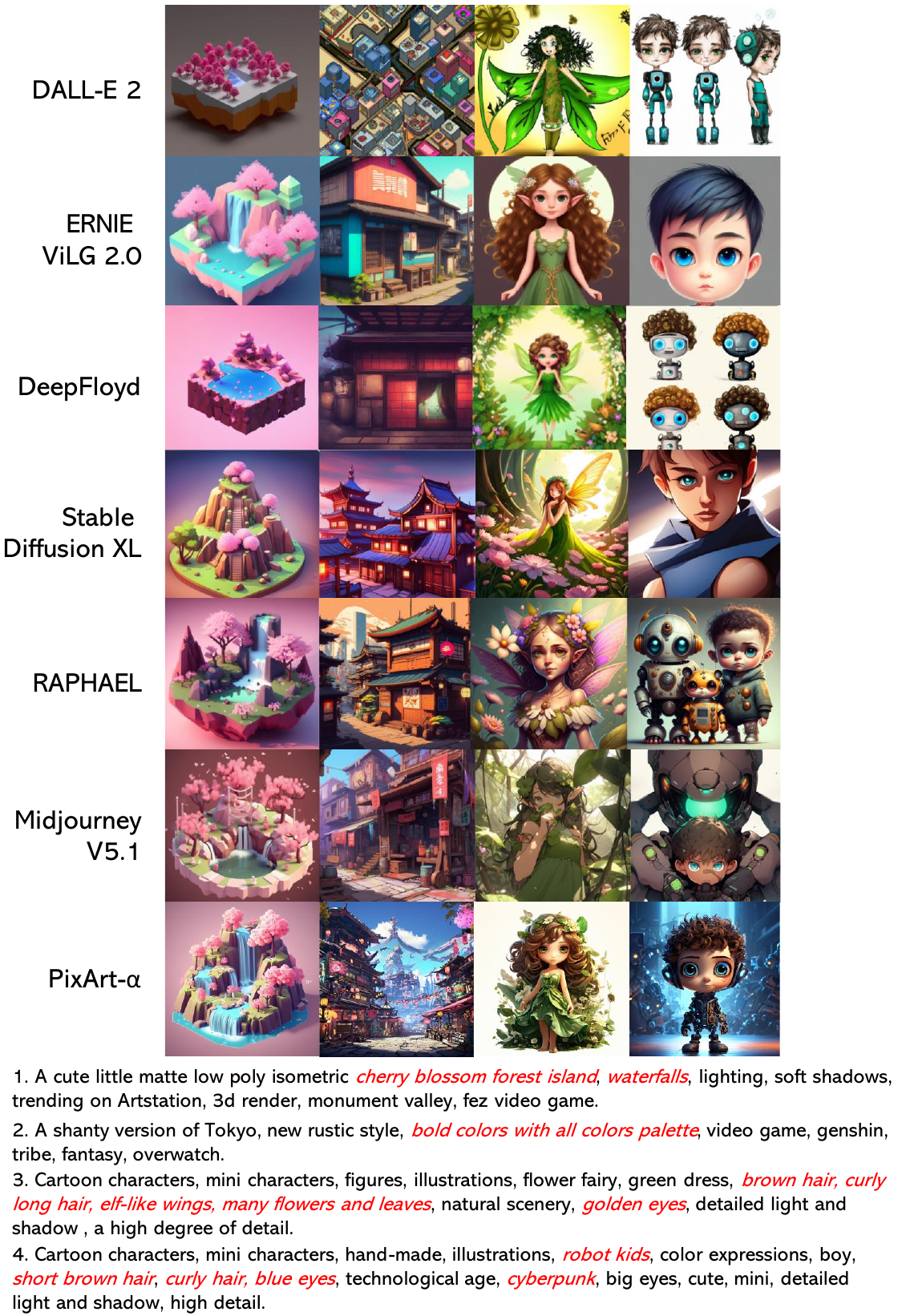}
\caption{The prompts~\citep{xue2023raphael} for each column are given in the figure. We give the comparisons between DALL-E 2 Midjourney v5.1, Stable Diffusion XL, ERNIE ViLG 2.0, DeepFloyd, and RAPHAEL. They are given the same prompts, where the words that the human artists yearn to preserve within the generated images are highlighted in red. Better zoom in 200\%.
}
\label{fig:comparefig_multiple_methods_more}
\end{center} 
\end{figure}

In Figure~\ref{fig:comparefig_multiple_methods} and~\ref{fig:comparefig_multiple_methods_more}, we present the comparison results using a test prompt selected by RAPHAEL. The instances depicted here exhibit performance that is on par with, or even surpasses, that of existing powerful generative models.

\subsection{Auto-labeling Techniques}
\begin{figure}[ht]
\begin{center}
\footnotesize
\includegraphics[width=1\linewidth]{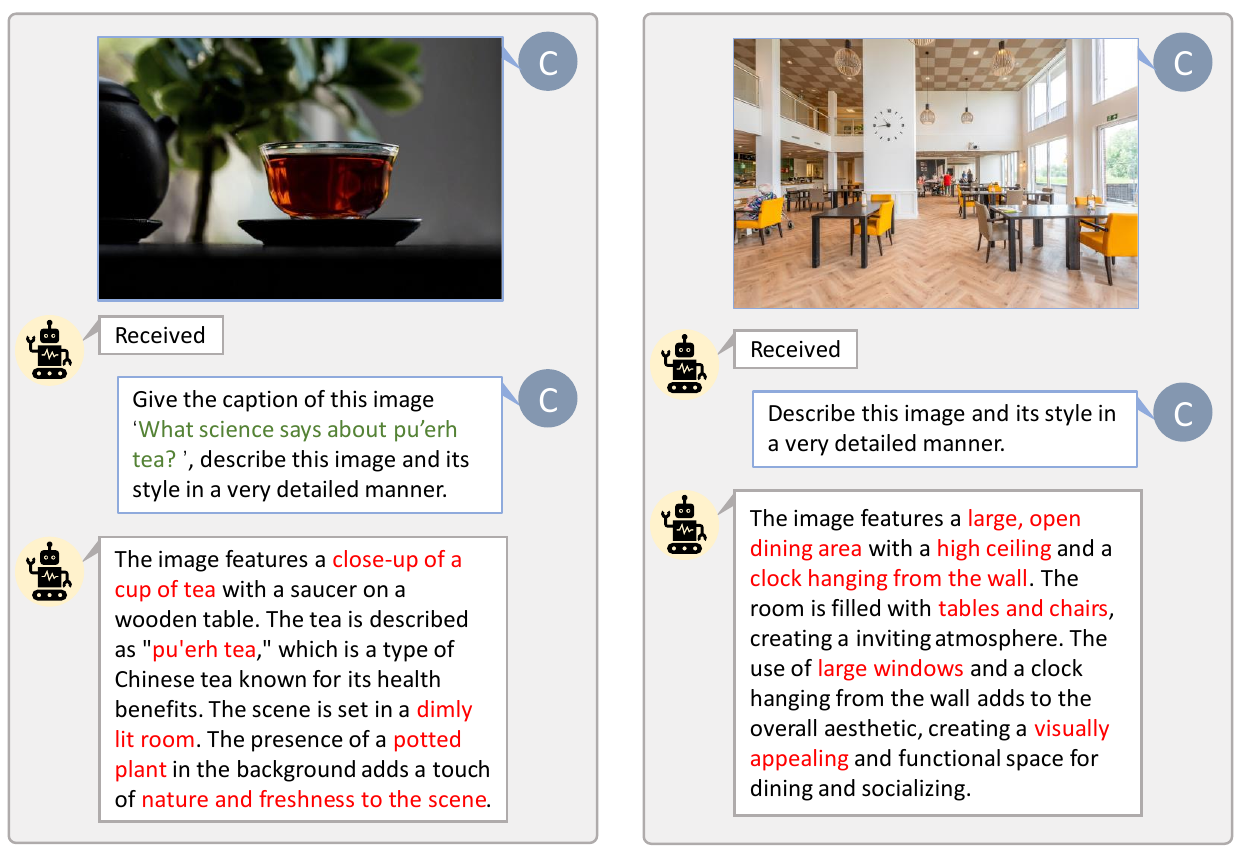}
\caption{We present auto-labeling with custom prompts for LAION~(left) and SAM~(right). The words highlighted in green represent the original caption in LAION, while those marked in red indicate the detailed captions labeled by LLaVA.
}
\label{fig:llava-dialog}
\end{center} 
\end{figure}

\begin{figure}[ht]
    \centering
    \includegraphics[width=\textwidth]{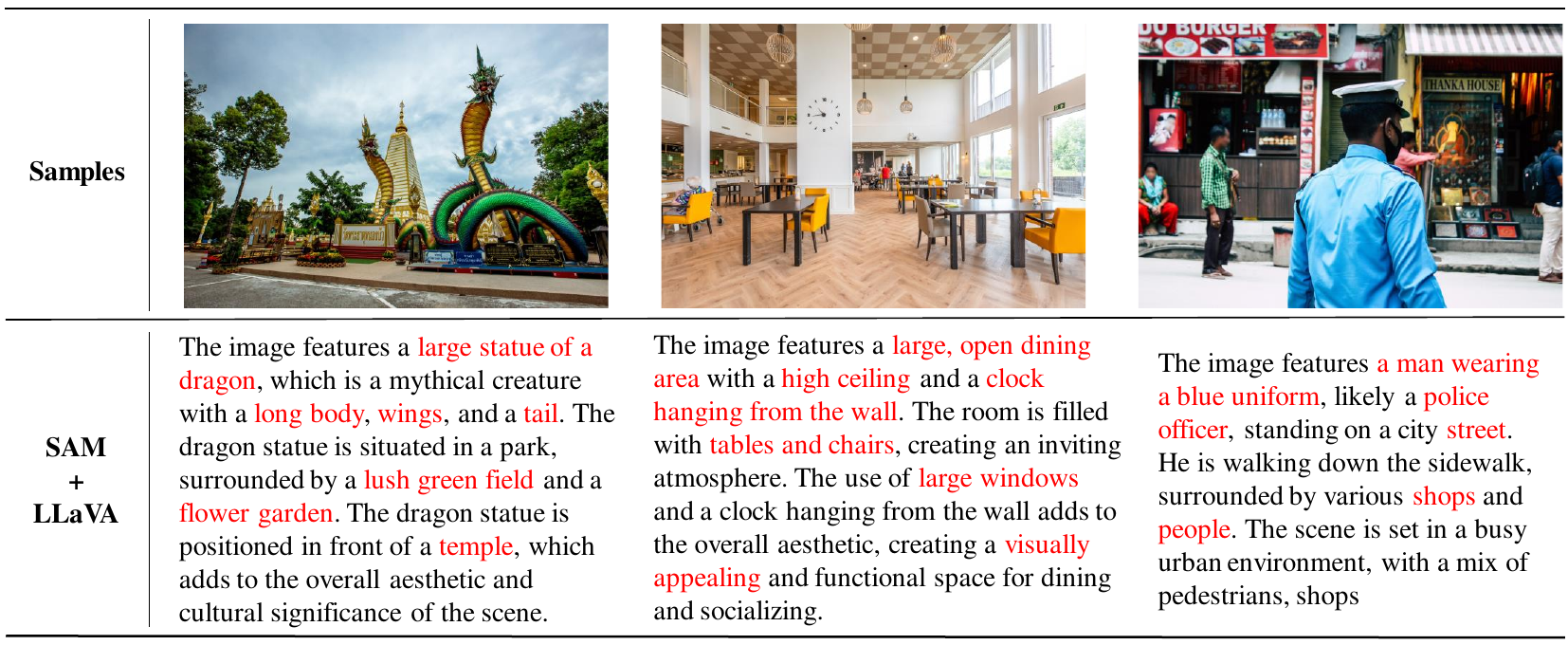}
    \caption{Examples from the SAM dataset using LLaVA-produced labels. The detailed image descriptions in LLaVA captions can aid the model to grasp more concepts per iteration and boost text-image alignment efficiency.}
    \label{fig:text-image-samples-sam}
\vspace{-1mm}
\end{figure}

To generate captions with high information density, we leverage state-of-the-art vision-language models LLaVA~\citep{liu2023llava}. Employing the prompt, ``\textit{Describe this image and its style in a very detailed manner}'', we have significantly improved the quality of captions. We show the prompt design and process of auto-labeling in Figure~\ref{fig:llava-dialog}. More image-text pair samples on the SAM dataset are shown in Figure~\ref{fig:text-image-samples-sam}.

\subsection{Additional Implementation Details}\label{sec:pixel}
We include detailed information about all of our \model{} models in this section. As shown in Table~\ref{tab:additional_implementation}, among the 256$\times$256 phases, our model primarily focuses on the text-to-image alignment stage, with less time on fine-tuning and only 1/8 of that time spent on ImageNet pixel dependency.
\paragraph{\model{} model details.} For the embedding of input timesteps, we employ a 256-dimensional frequency embedding~\citep{dhariwal2021diffusionbeatgan}. This is followed by a two-layer MLP that features a dimensionality matching the transformer's hidden size, coupled with SiLU activations. We adopt the DiT-XL model, which has 28 Transformer blocks in total for better performance, and the patch size of the PatchEmbed layer in ViT~\citep{dosovitskiy2020vit} is 2$\times$.
\paragraph{Multi-scale training.} Inspired by \cite{podell2023sdxl}, we incorporate the multi-scale training strategy into our pipeline. Specifically, We divide the image size into 40 buckets with different aspect ratios, each with varying aspect ratios ranging from 0.25 to 4, mirroring the method used in SDXL. During optimization, a training batch is composed using images from a single bucket, and we alternate the bucket sizes for each training step. In practice, we only apply multi-scale training in the high-aesthetics stage after pretraining the model at a fixed aspect ratio and resolution~(\ie 256px).
We adopt the positional encoding trick in DiffFit~\citep{xie2023difffit} since the image resolution and aspect change during different training stages. 
\hl{\paragraph{Additional time consumption.} Beside the training time discussed in Table~\ref{tab:additional_implementation}, data labeling and VAE training may need additional time. We treat the pre-trained VAE as a ready-made component of a model zoo, the same as pre-trained CLIP/T5-XXL text encoder, and our total training process does not include the training of VAE. However, our attempt to train a VAE resulted in an approximate training duration of 25 hours, utilizing 64 V100 GPUs on the OpenImage dataset. As for auto-labeling, we use LLAVA-7B to generate captions. LLaVA's annotation time on the SAM dataset is approximately 24 hours with 64 V100 GPUs. To ensure a fair comparison, we have temporarily excluded the training time and data quantity of VAE training, T5 training time, and LLaVA auto-labeling time.}

\paragraph{Sampling algorithm.} In this study, we incorporated three sampling algorithms, namely iDDPM~\citep{nichol2021iddpm}, DPM-Solver~\citep{lu2022dpms}, and SA-Solver~\citep{xue2023sas}. We observe these three algorithms perform similarly in terms of semantic control, albeit with minor differences in sampling frequency and color representation. To optimize computational efficiency, we ultimately chose to employ the DPM-Solver with 20 inference steps.

\begin{table}[ht]
    \centering
    \caption{We report detailed information about every \model~training stage in our paper. Note that HQ~(High Quality) dataset here includes 4M JourneyDB~\citep{pan2023journeydb} and 10M internal data. The count of GPU days excludes the time for VAE feature extraction and T5 text feature extraction, as we offline prepare both features in advance so that they are not part of the training process and contribute no extra time to it.}
    \label{tab:additional_implementation}
    \resizebox{1.\linewidth}{!}{ 
    \begin{tabular}{l|ccccccc}
    \toprule
    Method & Stage & Image Resolution & $\mathrm{\#}$Images & Training Steps (K) & Batch Size & Learning Rate & GPU days (V100) \\
    \midrule
    \model & Pixel dependency  & 256$\times$256 & 1M ImageNet  & 300 & \hl{128$\times$8} & 2$\times{10^{-5}}$ & 88 \\
    \model & Text-Image align  & 256$\times$256 & 10M SAM    & \hl{150} & 178$\times$64 & 2$\times{10^{-5}}$ & 672 \\
    \model & High aesthetics      & 256$\times$256 & 14M HQ & \hl{90} & 178$\times$64 & 2$\times{10^{-5}}$ & 416 \\
    \midrule
    \model & High aesthetics      & 512$\times$512 & 14M HQ & 100 & 40$\times$64 & 2$\times{10^{-5}}$ & 320 \\
    \model & High aesthetics      & 1024$\times$1024 & 14M HQ & 16 & 12$\times$32 & 2$\times{10^{-5}}$ & 160 \\
    \bottomrule
    \end{tabular}
    }
\vspace{0pt}
\end{table}

\subsection{Hyper-parameters analysis}
In Figure~\ref{hyper-parameter}, we illustrate the variations in the model's metrics under different configurations across various datasets. we first investigate FID for the model and plot FID-vs-CLIP curves in Figure~\ref{fig:clip_fid} for 10k text-image paed from MSCOCO. The results show a marginal enhancement over SDv1.5. In Figure~\ref{fig:t2icompbench-bvqa} and~\ref{fig:t2icompbench-other}, we demonstrate the corresponding T2ICompBench scores across a range of classifier-free guidance~(cfg)~\citep{ho2022classifierfree} scales. The outcomes reveal a consistent and commendable model performance under these varying scales.

\subsection{More Images generated by \model}
\begin{figure}[ht]
\begin{center}
\footnotesize
\includegraphics[width=1\linewidth]{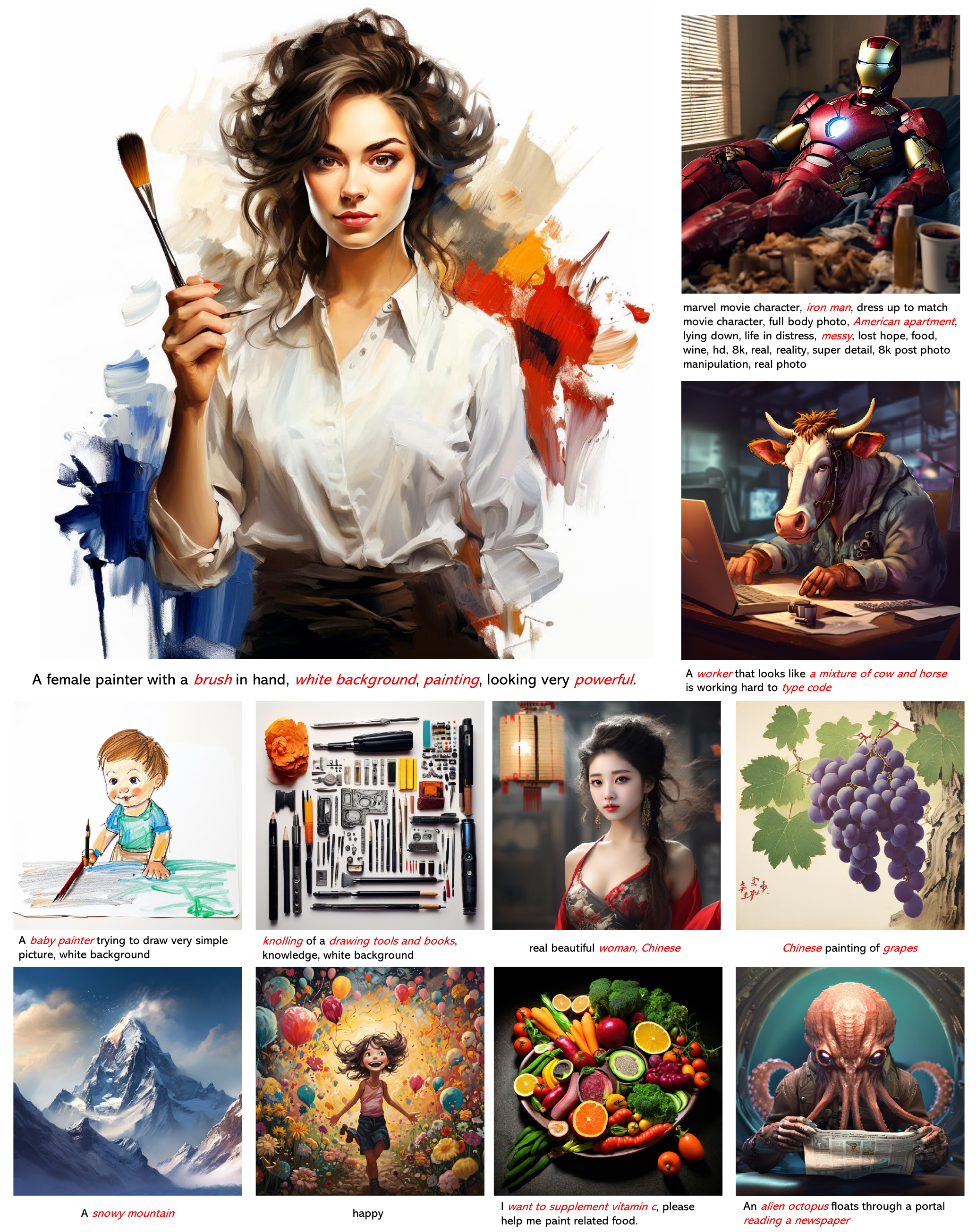}
\caption{The samples generated by \model{} demonstrate outstanding quality, marked by an exceptional level of fidelity and precision in aligning with the given textual descriptions. Better zoom in 200\%.}
\label{fig:appendix_more_samples}
\end{center} 
\end{figure}

\begin{figure}[ht]
\begin{center}
\footnotesize
\includegraphics[width=1\linewidth]{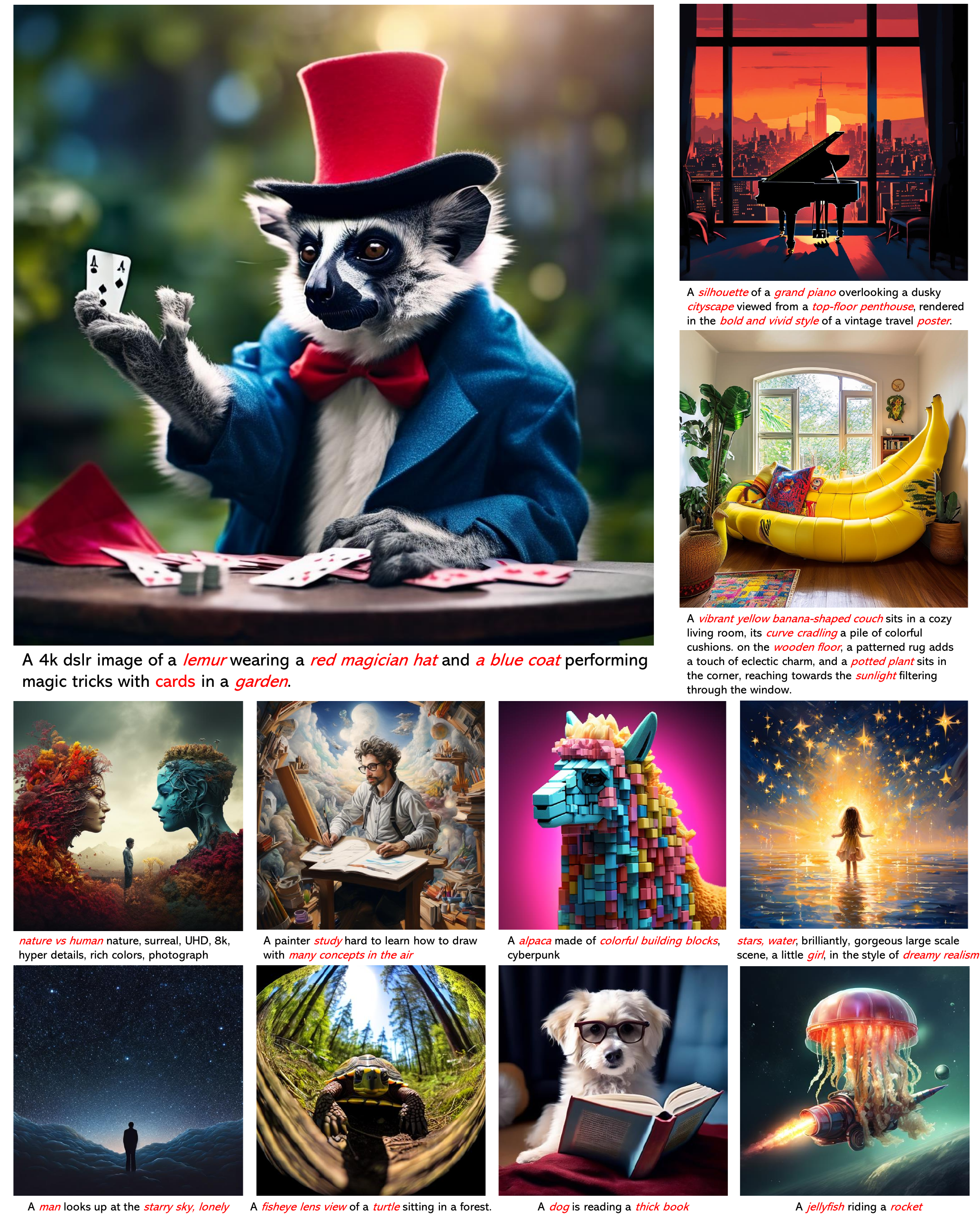}
\caption{The samples generated by \model{} demonstrate outstanding quality, marked by an exceptional level of fidelity and precision in aligning with the given textual descriptions. Better zoom in 200\%.}
\label{fig:appendix_more_samples2}
\end{center} 
\end{figure}

\begin{figure}[ht]
\begin{center}
\footnotesize
\includegraphics[width=1\linewidth]{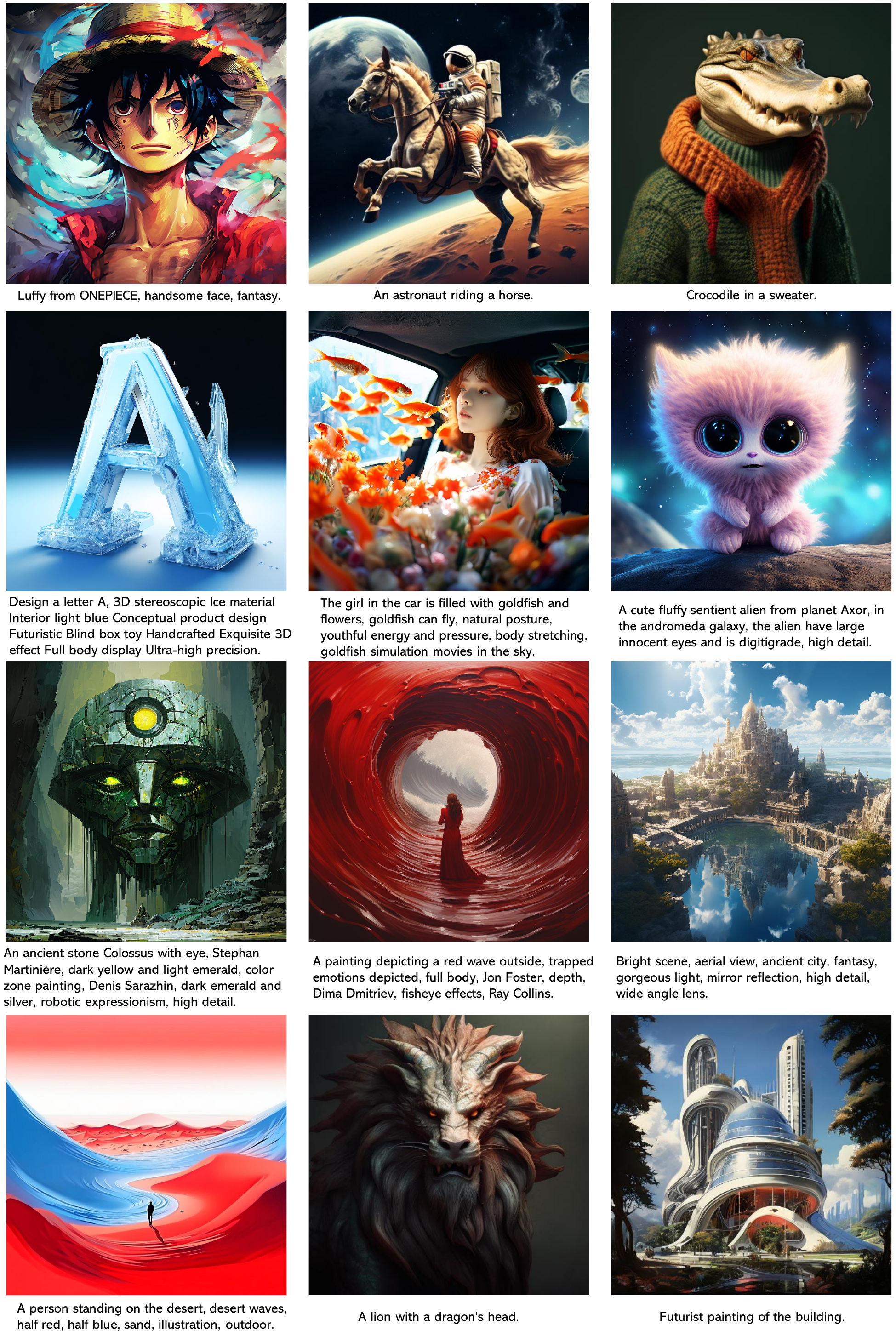}
\caption{The samples generated by \model{} demonstrate outstanding quality, marked by an exceptional level of fidelity and precision in aligning with the given textual descriptions. Better zoom in 200\%.
}
\label{fig:appendix_more_samples3}
\end{center} 
\end{figure}

\begin{figure}[ht]
\begin{center}
\footnotesize
\includegraphics[width=1.0\linewidth]{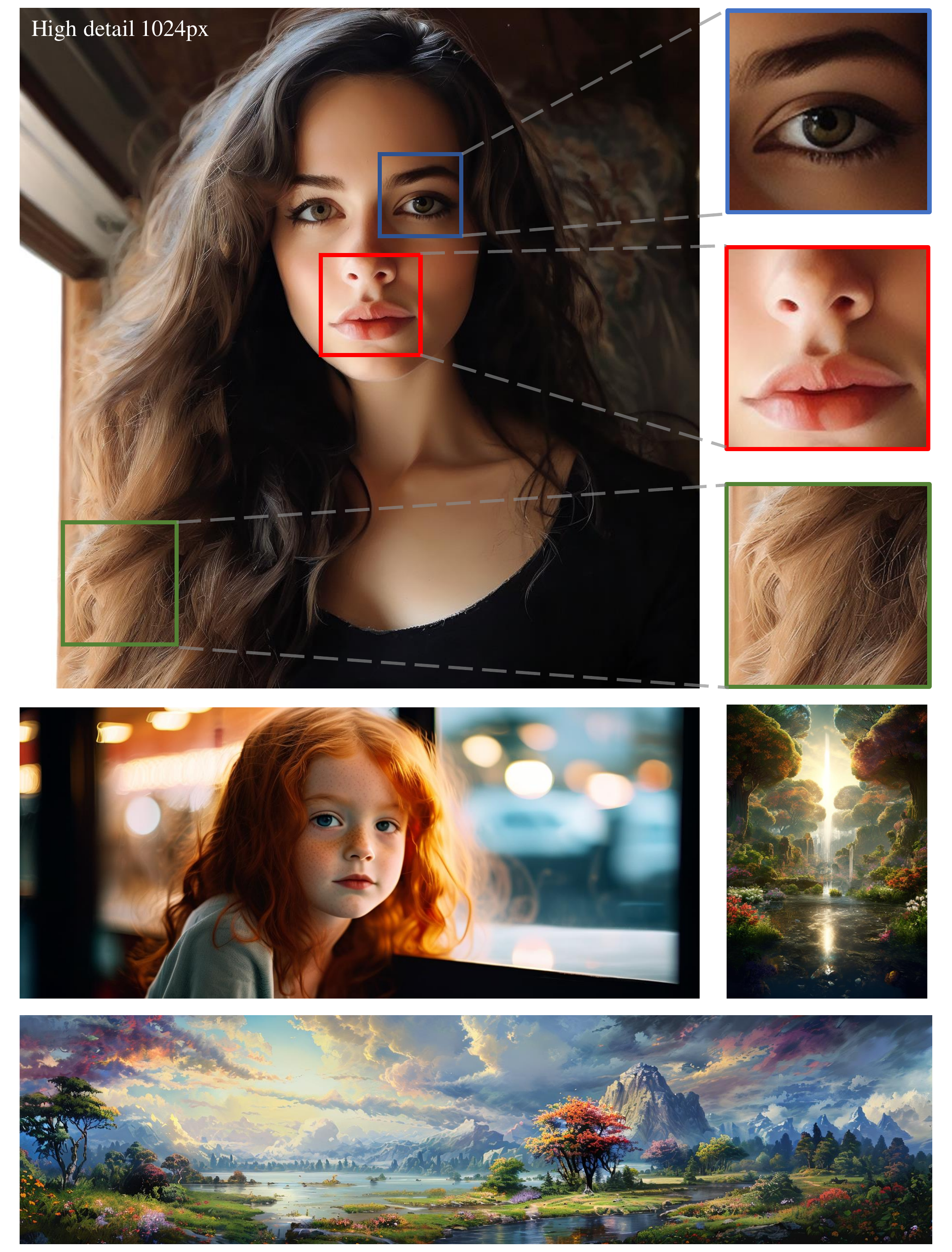}
\caption{\model{} is capable of generating images with resolutions of up to $1024\times1024$ while preserving rich, complex details. Additionally, it can generate images with arbitrary aspect ratios, providing flexibility in image generation.
}
\label{fig:appendix-more-samples-highresolution}
\end{center} 
\end{figure}

More visual results generated by \model{} are shown in Figure~\ref{fig:appendix_more_samples}, ~\ref{fig:appendix_more_samples2}, and~\ref{fig:appendix_more_samples3}. The samples generated by \model{} demonstrate outstanding quality, marked by their exceptional fidelity and precision in faithfully adhering to the given textual descriptions. As depicted in Figure~\ref{fig:appendix-more-samples-highresolution}, \model{} demonstrates the ability to synthesize high-resolution images up to $1024 \times 1024$ pixels and contains rich details, and is capable of generating images with arbitrary aspect ratios, enhancing its versatility for real-world applications. Figure~\ref{fig:appendix-more-samples-styletransf} illustrates \model{}'s remarkable capacity to manipulate image styles through text prompts directly, demonstrating its versatility and creativity.

\definecolor{green}{RGB}{208, 232, 196}
\definecolor{blue}{RGB}{180, 199, 231}

\begin{figure}[ht]
\begin{center}
\footnotesize
\includegraphics[width=1.0\linewidth]{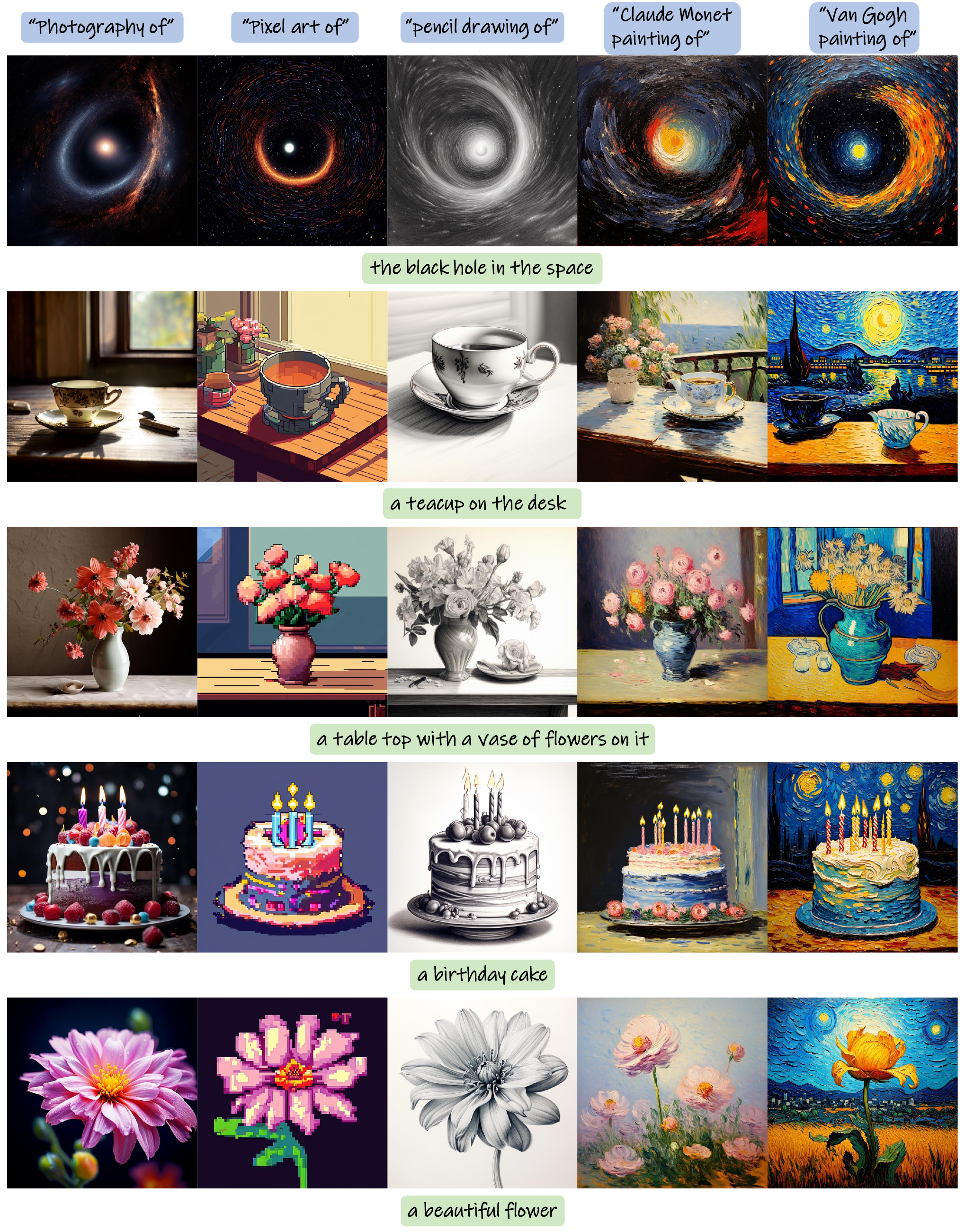}
\caption{\textbf{Prompt mixing:} \model{} can directly manipulate the image style with text prompts. In this figure, we generate five outputs using the \colorbox{blue}{styles} to control \colorbox{green}{the objects}. For instance, the second picture of the first sample, located at the left corner of the figure, uses the prompt ``\colorbox{blue}{Pixel Art of} \colorbox{green}{the black hole in the space}''. Better zoom in 200\%.}
\label{fig:appendix-more-samples-styletransf}
\end{center} 
\end{figure}

\subsection{Disccusion of FID metric for evaluating image quality}\label{appendix:fid}
During our experiments, we observed that the FID~(Fréchet Inception Distance) score may not accurately reflect the visual quality of generated images. Recent studies such as SDXL~\citep{podell2023sdxl} and Pick-a-pic~\citep{kirstain2023pick} have presented evidence suggesting that the COCO zero-shot FID is negatively correlated with visual aesthetics. 

Furthermore, it has been stated by Betzalel et al.~\citep{betzalel2022study} that the feature extraction network used in FID is pretrained on the ImageNet dataset, which exhibits limited overlap with the current text-to-image generation data. Consequently, FID may not be an appropriate metric for evaluating the generative performance of such models, and ~\citep{betzalel2022study} recommended employing human evaluators for more suitable assessments.

Thus, we conducted a user study to validate the effectiveness of our method.

\label{sec:more_customization}
\subsection{Customized Extension}
In text-to-image generation, the ability to customize generated outputs to a specific style or condition is a crucial application. We extend the capabilities of \model by incorporating two commonly used customization methods: DreamBooth~\citep{ruiz2022dreambooth} and ControlNet~\citep{zhang2023adding}.

\paragraph{DreamBooth.} DreamBooth can be seamlessly applied to \model{} without further modifications. The process entails fine-tuning \model{} using a learning rate of 5e-6 for 300 steps, without the incorporation of a class-preservation loss.

\begin{figure}[!ht]
    \centering
    \begin{subfigure}[b]{\textwidth}
        \includegraphics[width=\textwidth]{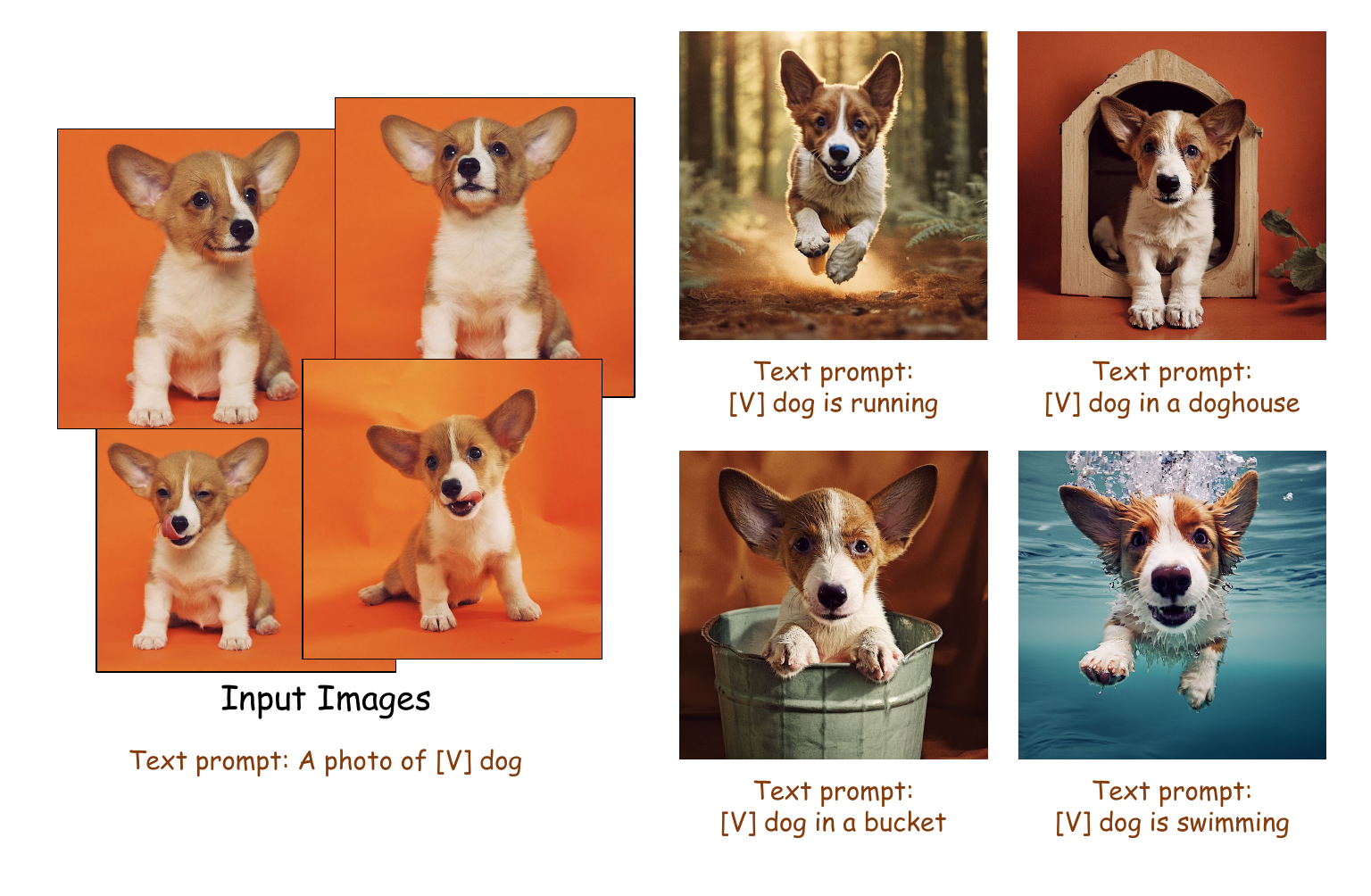}
        \caption{Dreambooth + \model{} is capable of customized image generation aligned with text prompts.}
        \label{fig:dreambooth_dog}
    \end{subfigure}
    \hfill
    \begin{subfigure}[b]{\textwidth}
        \includegraphics[width=\textwidth]{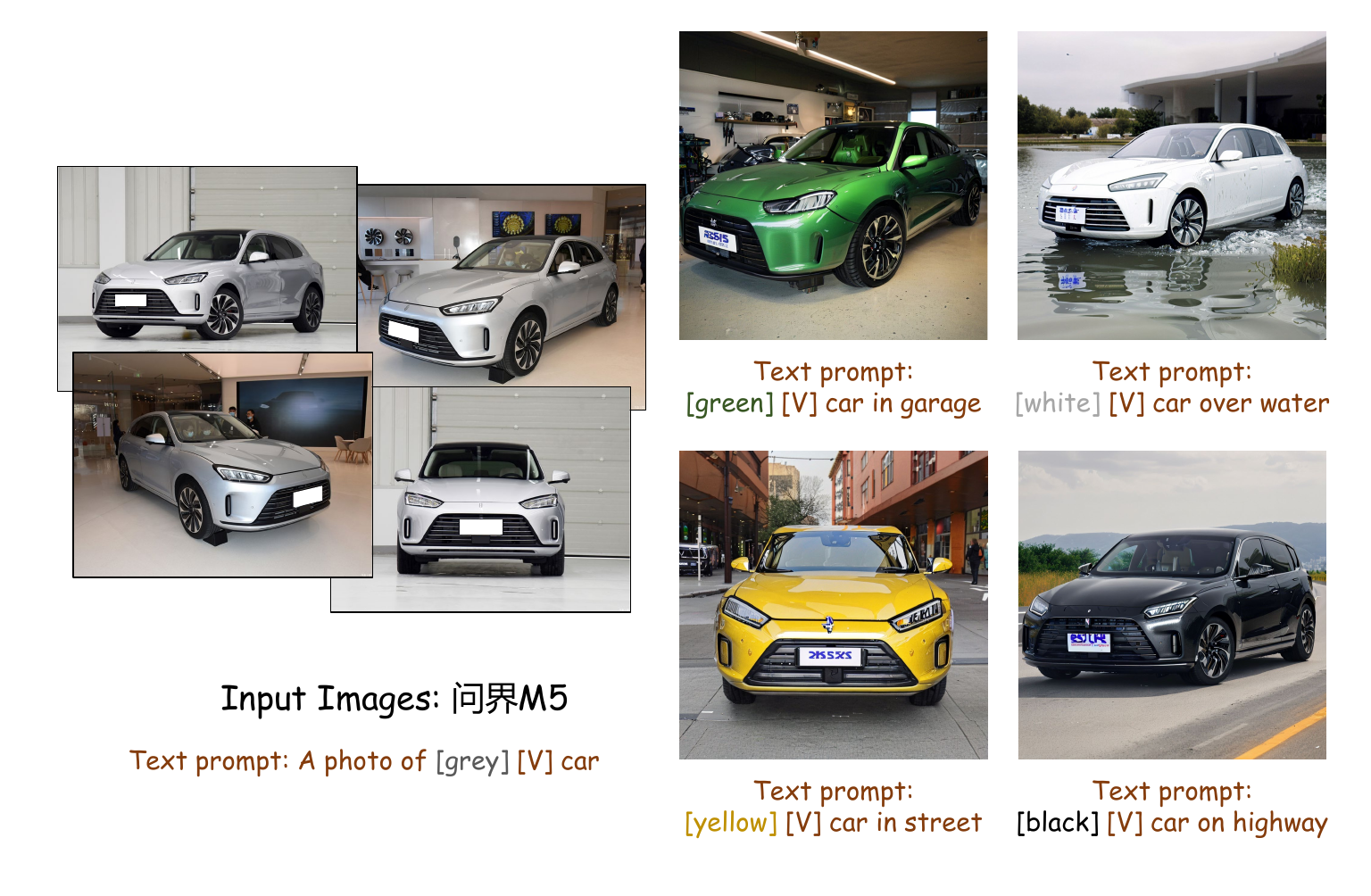}
        \caption{Dreambooth + \model{} is capable of color modification of a specific object such as Wenjie M5.}
        \label{fig:dreambooth_car}
    \end{subfigure}
    \caption{\model{} can be combined with Dreambooth. Given a few images and text prompts, \model{} can generate high-fidelity images, that exhibit natural interactions with the environment~\ref{fig:dreambooth_dog}, precise modification of the object colors~\ref{fig:dreambooth_car}, demonstrating that \model{} can generate images with exceptional quality, and has a strong capability in customized extension.
    }\label{fig:dreambooth}
\end{figure}

As depicted in Figure~\ref{fig:dreambooth_dog}, given a few images and text prompts, \model{} demonstrates the capacity to generate high-fidelity images. These images present natural interactions with the environment under various lighting conditions. Additionally, \model{} is also capable of precisely modifying the attribute of a specific object such as color, as shown in~\ref{fig:dreambooth_car}. Our appealing visual results demonstrate \model{} can generate images of exceptional quality and its strong capability for customized extension.

\paragraph{ControlNet.} Following the general design of ControlNet~\citep{zhang2023adding}, we freeze each DiT Block and create a trainable copy, augmenting with two zero linear layers before and after it. The control signal $c$ is obtained by applying the same VAE to the control image and is shared among all blocks. For each block, we process the control signal $c$ by first passing it through the first zero linear layer, adding it to the layer input $x$, and then feeding it into the trainable copy and the second zero linear layer. The processed control signal is then added to the output $y$ of the frozen block, which is obtained from input $x$. 
We trained the ControlNet on HED~\citep{xie2015holistically} signals using a learning rate of 5e-6 for 20,000 steps.

\begin{figure}[!ht]
    \centering
    \begin{subfigure}[b]{0.9\textwidth}
        \includegraphics[width=\textwidth]{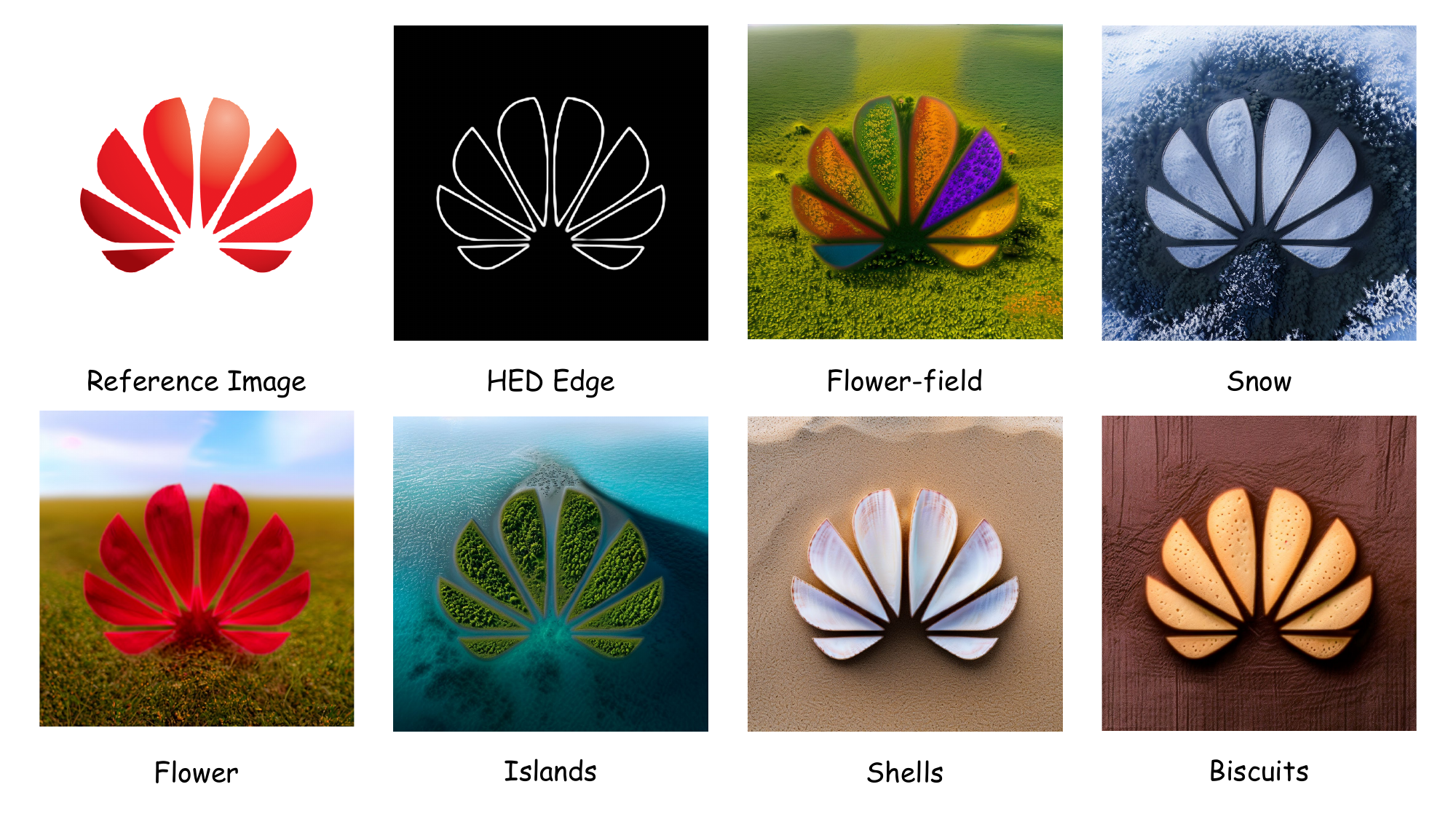}
    \end{subfigure}
    \begin{subfigure}[b]{0.9\textwidth}
        \includegraphics[width=\textwidth]{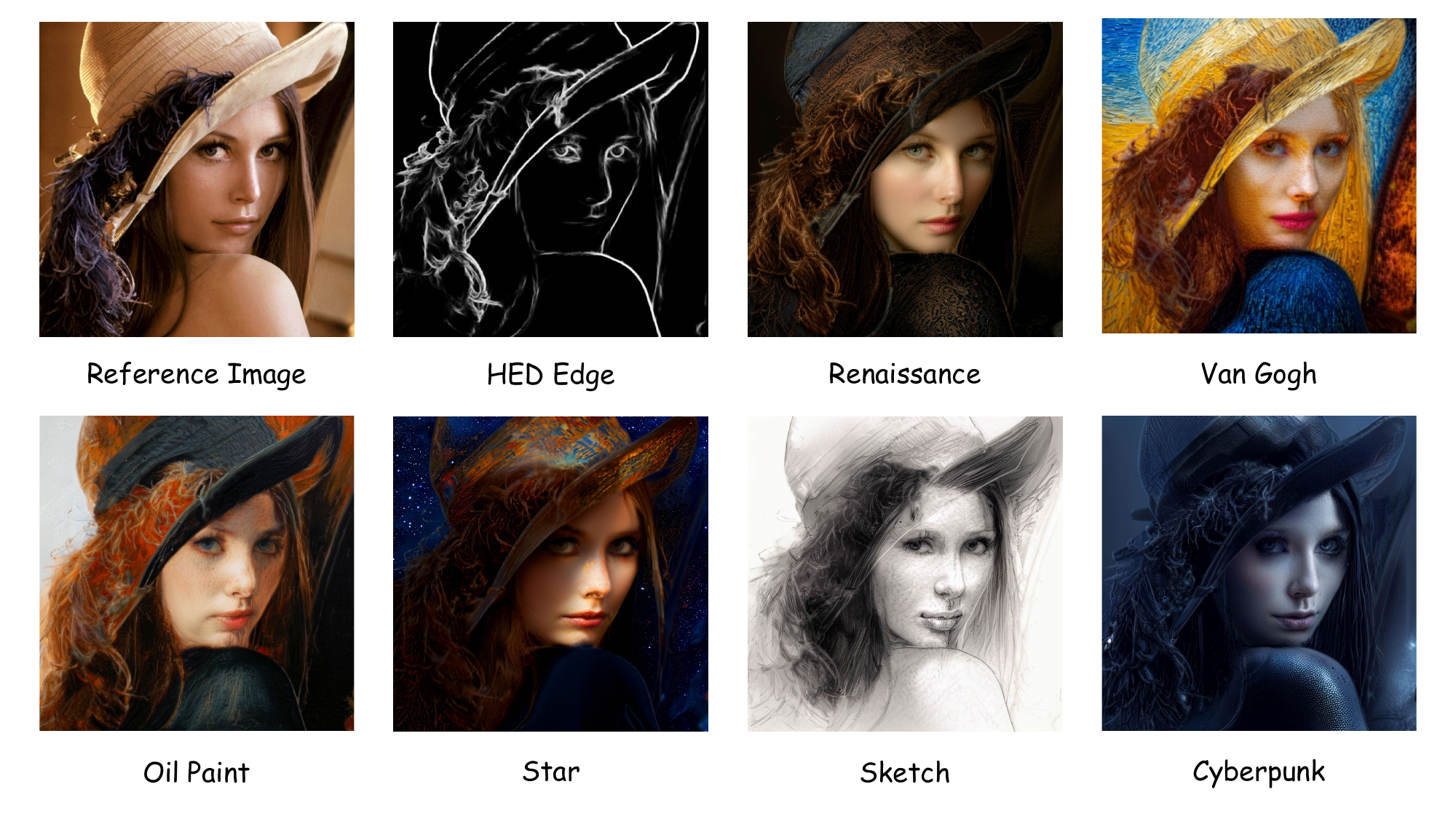}
    \end{subfigure}
    \caption{ControlNet customization samples from \model. We use the reference images to generate the corresponding HED edge images and use them as the control signal for \model ControlNet. Better zoom in 200\%.
    }\label{fig:ControlNet}
\end{figure}

As depicted in Figure~\ref{fig:ControlNet}, when provided with a reference image and control signals, such as edge maps, we leverage various text prompts to generate a wide range of high-fidelity and diverse images. Our results demonstrate the capacity of \model{} to yield personalized extensions of exceptional quality.


\subsection{Discussion on Transformer \vs U-Net}
The Transformer-based network's superiority over convolutional networks has been widely established in various studies, showcasing attributes such as robustness~\citep{zhou2022understanding,xie2021segformer}, effective modality fusion~\citep{girdhar2023imagebind}, and scalability~\citep{Peebles2022DiT}. Similarly, the findings on multi-modality fusion are consistent with our observations in this study compared to the CNN-based generator~(U-Net). For instance, Table~\ref{benchmark:t2icompbench} illustrates that our model, \model, significantly outperforms prevalent U-Net generators in terms of compositionality. This advantage is not solely due to the high-quality alignment achieved in the second training stage but also to the multi-head attention-based fusion mechanism, which excels at modeling long dependencies. This mechanism effectively integrates compositional semantic information, guiding the generation of vision latent vectors more efficiently and producing images that closely align with the input texts. These findings underscore the unique advantages of Transformer architectures in effectively fusing multi-modal information.

\subsection{Limitations \& Failure cases  }
\begin{figure}[ht]
\begin{center}
\footnotesize
\includegraphics[width=1\linewidth]{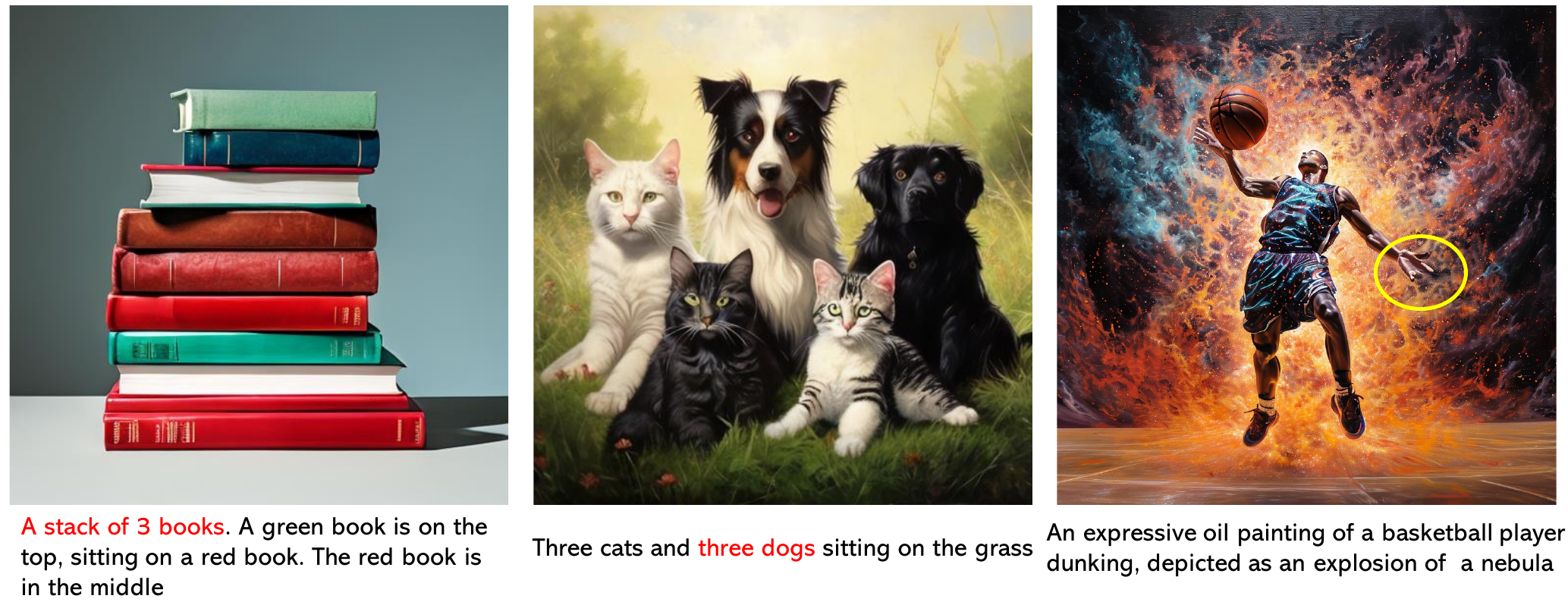}
\caption{Instances where \model{} encounters challenges include situations that necessitate precise counting or accurate representation of human limbs. In these cases, the model may face difficulties in providing accurate results.
}
\label{fig:failure_case}
\end{center} 
\end{figure}

In Figure~\ref{fig:failure_case}, we highlight the model's failure cases in red text and yellow circle. Our analysis reveals the model's weaknesses in accurately controlling the number of targets and handling specific details, such as features of human hands. Additionally, the model's text generation capability is somewhat weak due to our data's limited number of font and letter-related images. We aim to explore these unresolved issues in the generation field, enhancing the model's abilities in text generation, detail control, and quantity control in the future.

\pgfplotsset{
    every axis label/.append style={font=\small},
    every tick label/.append style={font=\tiny},
    width=1\textwidth, height=1\textwidth,
}

\begin{figure}
\centering
    \begin{subfigure}{0.33\textwidth}
    \begin{tikzpicture}
        \begin{axis}[
            ymin=12.2, ymax=23,
            xmin=0.24, xmax=0.267,
            xtick distance=0.01, 
            ytick distance=2, 
            grid=both,
            xlabel={CLIP score},
            ylabel={FID},
            legend columns=2,
            legend style={font=\fontsize{5pt}{3pt}\selectfont}, 
            legend pos=north east, 
        ]
        \addplot+[color=blue, thin] coordinates {(0.241, 22.4) (0.25,17.2) (0.258,15.2) (0.263,16) (0.264,17.5) (0.265,18.7)}; 
        \addlegendentry{SDv1.5}
        \addplot+[color=red, thin] coordinates {(0.249,14.61) (0.255,12.86) (0.2597,13.146) (0.2604,13.5685) (0.2588,15.17) (0.2549,20.5)};
        \addlegendentry{Our}
        \end{axis}
    \end{tikzpicture}
    \caption{FID on MSCOCO}
    \label{fig:clip_fid}
    \end{subfigure}\hfill
    \begin{subfigure}{0.33\textwidth}
    \begin{tikzpicture}
        \begin{axis}[
            ymin=0.52, ymax=0.79,
            xmin=1.5, xmax=7.5,
            xtick distance=1, 
            ytick distance=0.04, 
            grid=both,
            xlabel={cfg scale},
            ylabel={B-VQA},
            legend columns=2,
            legend style={font=\fontsize{5pt}{3pt}\selectfont}, 
            legend pos=north west, 
        ]
        \addplot+[color=blue, thin] coordinates {(2,0.6648) (3,0.6811) (4,0.6911) (4.5,0.6731) (5,0.6832)  (6,0.7019) (7,0.6913)};
        \addlegendentry{Color}
        \addplot+[color=red, thin] coordinates {(2,0.5207) (3,0.5260) (4,0.5418) (5,0.5441) (6,0.5530) (7,0.5588)};
        \addlegendentry{Shape}
        \addplot+[color=orange, mark=triangle*,thin] coordinates {(2,0.6524) (3,0.6609) (4,0.6800) (5,0.6939)  (6,0.7021) (7,0.6971)};
        \addlegendentry{Texture}
        \end{axis}
    \end{tikzpicture}
    \caption{T2i-CompBench}
    \label{fig:t2icompbench-bvqa}
    \end{subfigure}
    \hfill
    \begin{subfigure}{0.33\textwidth}
    \begin{tikzpicture}
        \begin{axis}[
            ymin=0.14, ymax=0.58,
            xmin=1.5, xmax=7.5,
            xtick distance=1, 
            ytick distance=0.1, 
            grid=both,
            xlabel={cfg scale},
            ylabel={UniDet  CLIP  3-in-1},
            legend columns=2,
            legend style={font=\fontsize{4pt}{3pt}\selectfont}, 
            legend pos=north west, 
        ]
        \addplot+[color=blue, thin] coordinates {(2,0.1775) (3,0.1865) (4,0.1919) (5,0.2130) (6, 0.2111) (7, 0.1933)};
        \addlegendentry{Spatial}
        \addplot+[color=red, thin] coordinates {(2,0.3158) (3,0.3160) (4,0.3165) (5,0.3184) (6, 0.3186) (7, 0.3188)};
        \addlegendentry{Non-spatial}
        \addplot+[color=orange, mark=triangle*, thin] coordinates {(2,0.4096) (3,0.4112) (4,0.4082) (5,0.4163) (6,0.4128) (7,0.4115)};
        \addlegendentry{Complex}
        \end{axis}
    \end{tikzpicture}
    \caption{T2i-CompBench}
    \label{fig:t2icompbench-other}
    \end{subfigure}
\caption{(a) Plotting FID \vs CLIP score for different cfg scales sampled from [1.5, 2.0, 3.0, 4.0, 5.0, 6.0]. \model{} shows slight better performance than SDv1.5 on MSCOCO. (b) and (c) demonstrate the ability of \model{} to maintain robustness across various cfg scales on the T2I-CompBench.}
\label{hyper-parameter}
\end{figure}
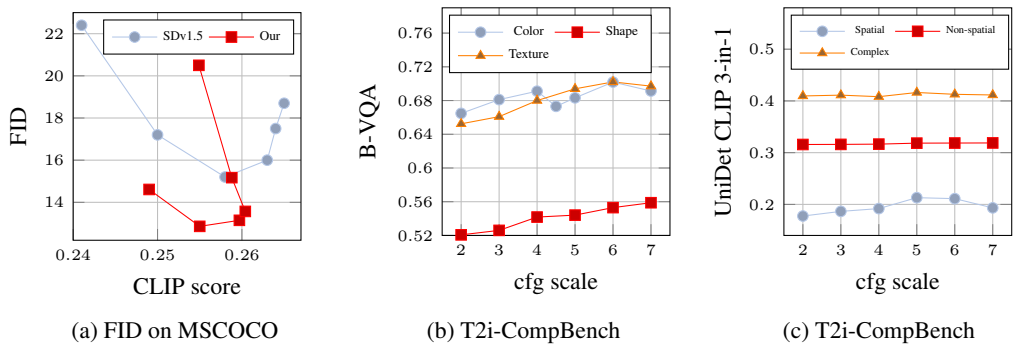

\subsection{Unveil the answer}
\begin{figure}[ht]
\begin{center}
\footnotesize
\includegraphics[width=1\linewidth]{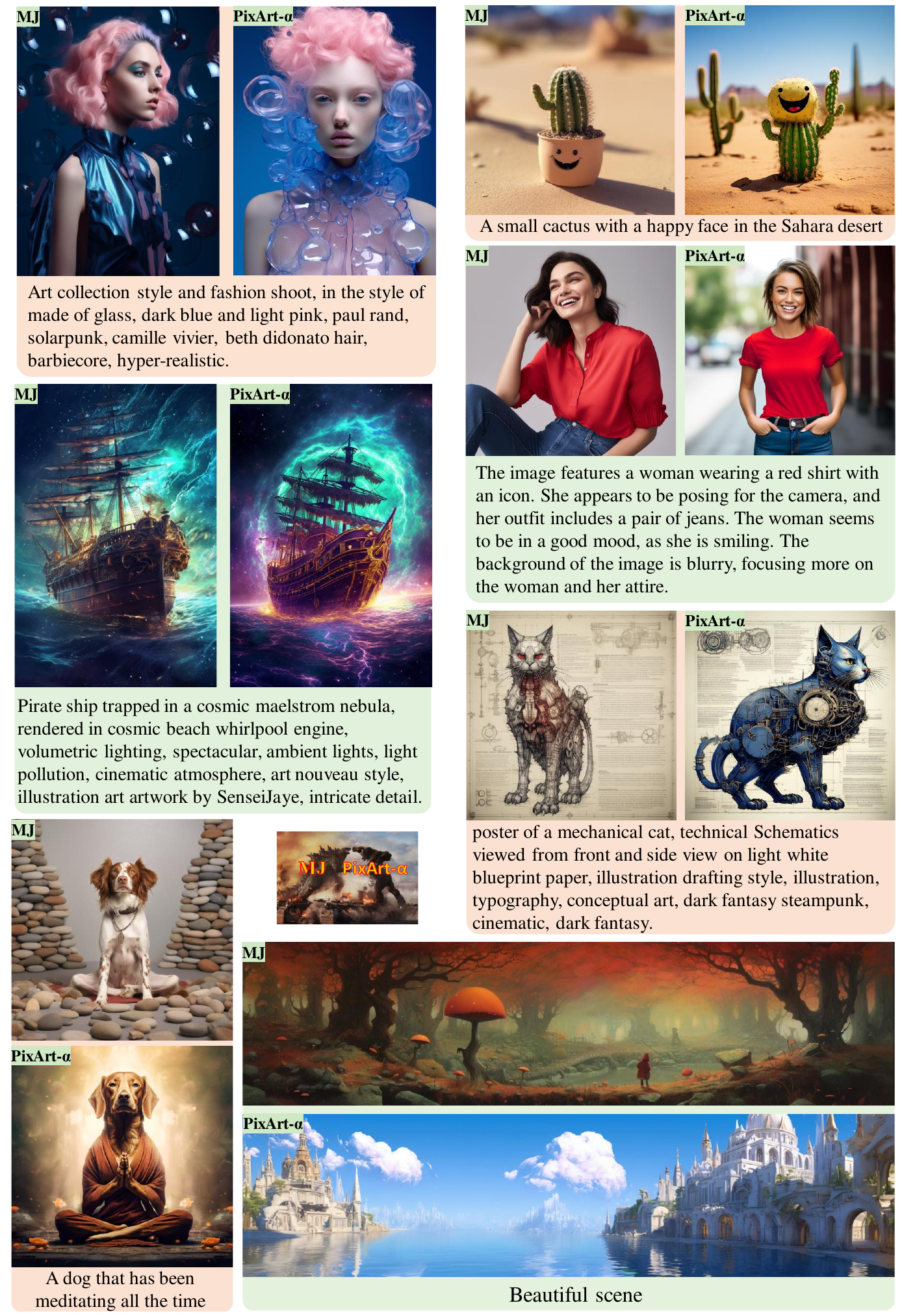}
\caption{This figure presents the \textbf{answers} to the image generation quality assessment as depicted in Appendix~\ref{sec:compare-mj}. The method utilized for each pair of images is annotated at the top-left corner. 
}
\label{fig:compare_mj_waterprint}
\end{center} 
\end{figure}

In Figure~\ref{fig:compare_mj}, we present a comparison between \model{} and Midjourney and conceal the correspondence between images and their respective methods, inviting the readers to guess. Finally, in Figure~\ref{fig:compare_mj_waterprint}, we unveil the answer to this question. It is difficult to distinguish between \model{} and Midjourney, which demonstrates \model{}'s exceptional performance.

\clearpage
\bibliography{iclr2024_conference}
\bibliographystyle{iclr2024_conference}

\end{document}